 \documentclass[acmtog]{acmart}
\acmSubmissionID{158}
\usepackage{booktabs} 
\usepackage{multirow}
\usepackage{tabularx}
\usepackage{wrapfig}
\usepackage{mathtools}
\usepackage{lipsum}
\usepackage{footmisc}
\newcommand\blfootnote[1]{%
  \begingroup
  \renewcommand\thefootnote{}\footnote{#1}%
  \addtocounter{footnote}{-1}%
  \endgroup
}

\usepackage[ruled]{algorithm2e} 

\SetAlFnt{\small}
\SetAlCapFnt{\small}
\SetAlCapNameFnt{\small}
\SetAlCapHSkip{0pt}

\begin{document}
\def\negativevspace{}	
\newcommand{\TODO}[1]{{\color{red}{[TODO: #1]}}}
\newcommand{\rh}[1]{{\color{green}#1}}
\newcommand{\dc}[1]{{\color{red}#1}}
\newcommand{\lzz}[1]{{\color{blue}#1}}
\newcommand{\phil}[1]{{\color[rgb]{0.2,0.8,0.2}#1}}
\newcommand{\jy}[1]{{\color[rgb]{0.7,0.2,0.0}#1}}
\newcommand{\edward}[1]{{\color[rgb]{0.7,0.2,0.7}#1}}
\newcommand{\ednote}[1]{{\color[rgb]{0.7,0.2,0.7}ED: #1}}
\newcommand{\para}[1]{\vspace{.05in}\noindent\textbf{#1}}
\newcommand{\xjqi}[1]{{\color{magenta}#1}}
\newcommand{\new}[1]{{\color{black}{#1}}}

\def\ie{\emph{i.e.}}
\def\eg{\emph{e.g.}}
\def\etal{{\em et al.}}
\def\etc{{\em etc.}}
\newcolumntype{C}[1]{>{\centering\arraybackslash}p{#1}}
	
\title{EXIM: A Hybrid Explicit-Implicit Representation for Text-Guided 3D Shape Generation}




\author{Zhengzhe Liu, Jingyu Hu, Ka-Hei Hui}
\affiliation{%
	\institution{The Chinese University of Hong Kong}\country{Hong Kong}}\email{zzliu,jyhu,khhui@cse.cuhk.edu.hk}
 \author{Xiaojuan Qi$^{\dagger}$}
\affiliation{%
	\institution{The University of Hong Kong}\country{Hong Kong}}\email{xjqi@eee.hku.hk}
\author{Daniel Cohen-Or}
\affiliation{%
	\institution{Tel-Aviv University}\country{Israel}}\email{cohenor@gmail.com}
\author{Chi-Wing Fu$^{\dagger}$}
\affiliation{%
	\institution{The Chinese University of Hong Kong}\country{Hong Kong}}\email{cwfu@cse.cuhk.edu.hk}
\renewcommand\shortauthors{Liu et al.}

\begin{teaserfigure}
\vspace*{-1mm}
  \centerline{\includegraphics[width=0.99\textwidth]{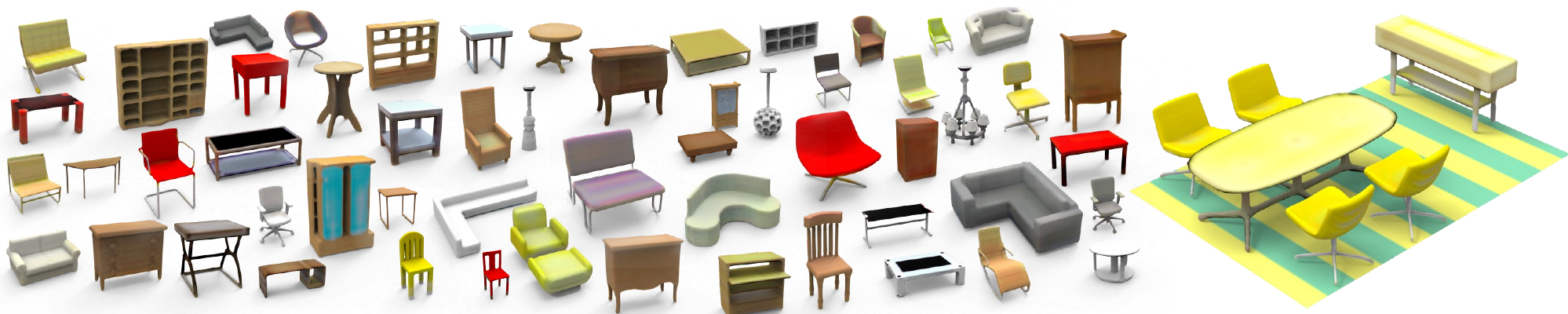}}
\vspace*{-3.5mm}
\caption{Left: a gallery of 3D shapes generated by our approach.
Right: a style-consistent indoor scene produced by our approach from a text description.
}
\label{fig:teaser}
\end{teaserfigure}

%
%


%
%



\setcopyright{acmlicensed}
\acmJournal{TOG}
\acmYear{2023} \acmVolume{42} \acmNumber{6} \acmArticle{228}
\acmMonth{12} \acmPrice{15.00}\acmDOI{10.1145/3618312}

\begin{abstract}
\blfootnote{$^\dagger$: Corresponding authors}

This paper presents a new text-guided technique for generating 3D shapes.
The technique leverages a hybrid  3D shape representation, \new{namely EXIM}, combining the strengths of explicit and implicit representations. Specifically, the explicit stage controls the topology of the generated 3D shapes and enables \new{local modifications}, \new{whereas} the implicit stage refines the shape and paints it with plausible colors. 
%
Also, the hybrid approach separates \new{the} shape and color and generates color conditioned on shape to ensure shape-color consistency. Unlike \new{the} existing state-of-the-art methods, we achieve high-fidelity shape generation from natural-language descriptions without the need for time-consuming per-shape optimization or reliance on human-annotated texts during training or test-time optimization.
Further, we demonstrate the \new{applicability} of our approach to generate indoor scenes with consistent styles using 
text-induced 3D shapes.
Through extensive experiments, we 
demonstrate the compelling quality of our results and 
the high coherency of our generated shapes with the input texts, surpassing the performance of existing methods by a significant margin. \new{Codes and models are released at 
\url{https://github.com/liuzhengzhe/EXIM}}.

\end{abstract}

\maketitle
\section{Introduction}
\label{sec:intro}

3D shape generation has a wide range of applications, including CAD, extended reality, games, and movies. To create 3D shapes, a highly natural and intuitive approach is to use natural language or texts. Text-guided 3D shape generation can improve the efficiency of 3D modeling, allowing non-experts to efficiently create shapes by providing a text description of the desired shape. 

To enable text-guided 3D shape generation, existing works~\cite{jain2021zero,mohammad2022clip,poole2022dreamfusion,lin2022magic3d} propose to leverage large-scale pre-trained vision-language models such as CLIP~\cite{radford2021learning} and text-to-image generation models~\cite{ramesh2022hierarchical,saharia2022photorealistic,rombach2022high} to optimize 3D shapes at test time. Yet, the optimization is time-consuming, \eg, 72 min.  for Dream Fields~\cite{jain2021zero} and 90 min. for Dream Fusion~\cite{poole2022dreamfusion} to generate a single 3D shape. 
Also, due to the lack of awareness of the 3D prior, their generative fidelity is still not satisfactory,~\eg, with unrealistic structures; see Figure~\ref{fig:swivel} in ``Stable-Dreamfusion'' \cite{poole2022dreamfusion}. 


Another {line} of works~\cite{chen2018text2shape,liu2022towards,sanghi2021clip,fu2022shapecrafter,wei2023taps3d,cheng2023sdfusion,jun2023shap} {aims} to generate shapes without test-time optimization. By learning the 3D prior from a shape collection,~\eg, ShapeNet~\cite{shapenet2015}, shapes can be generated with great efficiency in a feed-forward manner.
So, these approaches particularly suit generating shapes in specific classes, such as room assets.

{However, they still face several challenges. One major challenge arises from the need for appropriate 3D representations for text-guided 3D generation.}
Some works adopt explicit representations,~\eg, voxel grids~\cite{chen2018text2shape,mittal2022autosdf} and point clouds~\cite{nichol2022point}; yet, they cannot represent fine details due to their representation capabilities (See Figure~\ref{fig:swivel}: Point-E \cite{nichol2022point}). Others utilize implicit neural representations~\cite{liu2022towards,sanghi2021clip,fu2022shapecrafter} to first learn a latent feature vector/grid of the shape, then train an MLP to predict the occupancy  of the query point. However, their generative quality is still far from satisfactory,~\eg, with incomplete structures (see Figure~\ref{fig:swivel}: TITG3SG~\cite{liu2022towards}, CLIP-Forge~\cite{sanghi2021clip}, and ShapeCrafter \cite{fu2022shapecrafter}). 
{Moreover, the implicit representations used in these approaches encode global (non-local) features during the encoding process. As a result, achieving spatial disentanglement becomes challenging, making it difficult to  perform shape manipulation  
while leaving other regions unaffected. Further details will be presented in Section~\ref{sec:experiments}.}
%

Color is an {important} element in text-to-shape generation; yet, simply adopting existing 3D shape representations like voxel grid~\cite{mittal2022autosdf} and implicit fields~\cite{sanghi2021clip,fu2022shapecrafter} cannot handle color generation. An early work~\cite{chen2018text2shape} attempts to represent colors with voxels; its coarse results demonstrate that explicit representation like voxels is not a good choice for color representation. 
In more recent developments, researchers have explored the utilization of a joint shape-color implicit representation \cite{liu2022towards} or adopt models for unconditional 3D shape generation like GET3D \cite{gao2022get3d}.
%
However, their generative qualities are still not satisfying; see Figure~\ref{fig:swivel} ``TITG3SG'' and ``TAPS3D''. More importantly, the color of 3D shapes should be conditioned on the shape to ensure shape-color consistency; yet, in their methods, shape and color are  predicted in parallel, so it is challenging to manipulate one while keeping the other consistent.




In this work, we present a new representation called EXIM, a hybrid explicit-implicit representation for high-fidelity text-guided 3D shape generation, \new{enabling a certain level of local modification}. It is a two-stage coarse-to-fine pipeline.  First, we adopt the explicit compact voxel representation \new{inspired by~\cite{hui2022neural}} to create a coarse shape by predicting TSDF values, then locally condition on the stage-1 output to enhance the shape details and also to produce colors based on the text descriptions using neural implicit representations.
As Figure~\ref{fig:teaser} shows, our design is capable of generating high-fidelity 3D shapes from textual descriptions. 


Further, EXIM is flexible.
First, our explicit representation is spatially disentangled, allowing \new{the operations to focus on the
local regions of interest, while avoiding the other regions.}  
Second, EXIM decomposes shape and color generation, in which the color prediction is conditioned on the shape. This design encourages shape-color consistency when modifying the shape and color, while enabling independent modification on shape and color.

Lastly, to enrich the existing text-to-shape datasets, we present a  study utilizing pre-trained vision-language foundation models, such as BLIP \cite{li2023blip}, to generate textual descriptions for 3D shapes based on rendered images.  This enables us to effortlessly expand the existing dataset, incorporating new categories without the need for extensive human annotations. 

%


To evaluate our approach, we conduct experiments on multiple categories of ShapeNet~\cite{shapenet2015}. Both qualitative and quantitative results demonstrate that  EXIM outperforms existing works by a large margin in terms of generative fidelity and modification capability. Further, our approach can create style-consistent and plausible indoor scenes through generating shapes using text descriptions, as \new{demonstrated} 
in Figure~\ref{fig:teaser}.


\section{Related Work}
\label{sec:rw}


\paragraph{3D Shape Representation and Generation}

There are many ways to represent a 3D shape, 
such as point clouds~\cite{li2021sp,hui2020progressive,vahdat2022lion}, voxels~\cite{wu2016learning,smith2017improved}, meshes~\cite{tang2019skeleton}, and neural implicit functions~\cite{mescheder2019occupancy,chen2019learning,shue20223d,zhang20233dshape2vecset}.
Among the representations, the voxel-based explicit representation effectively captures \new{the} shape topology, but high-resolution voxels require substantial memory. To address this issue,~\cite{hui2022neural} proposes compressing \new{the} SDF volumes into a compact wavelet representation to retain more details by representing shapes in a compressed domain. 
Neural implicit functions offer a highly compact representation and excel at capturing surface details~\cite{chibane2020implicit,mittal2022autosdf,yan2022shapeformer,zhang20223dilg} and colors~\cite{oechsle2019texture,niemeyer2020differentiable}. In this work, we propose \new{EXIM}, a novel hybrid explicit-implicit representation for text-guided 3D shape generation, unifying the strengths of explicit and implicit representations. Our new 3D shape representation achieves high generative fidelity and supports \new{a certain level of local shape and color modification, as demonstrated in the various results shown in Section}~\ref{sec:experiments}.

\paragraph{Text-Guided 3D Shape Generation}

Existing works on text-guided 3D shape generation can be broadly categorized into two branches: optimizing the shape at test time~\cite{jain2021zero,michel2021text2mesh,chen2022tango,mohammad2022clip,liu2023iss,poole2022dreamfusion,lin2022magic3d,chen2023fantasia3d} and generating in a feed-forward manner~\cite{chen2018text2shape,mittal2022autosdf,liu2022towards,sanghi2021clip,fu2022shapecrafter,sanghi2022textcraft,li2022diffusion,wei2023taps3d,cheng2023sdfusion,jun2023shap}. Although the former branch covers a relatively wide range of generative categories, the latter branch generally offers better generative fidelity, efficiency, and stability. Consequently, we choose to explore this direction to develop our method.
As discussed in Section~\ref{sec:intro}, existing works in the latter branch still exhibit various limitations, such as inferior generative fidelity and limited capability of modifying the generated shapes. Besides, some existing works focus on text-to-NeRF (Neural Radiance Fields)~\cite{jiang20223d,wang2021clipnerf,li20223ddesigner} and avatar generation~\cite{hong2022avatarclip}. Unlike the objective of this work, they primarily concentrate on \new{the} image rendering qualities using NeRF rather than shape generation quality.

\paragraph{Text-Guided 3D Shape Modification}

Compared to text-guided 3D shape generation, text-guided 3D shape modification has been relatively less explored. \cite{liu2022towards} proposes a two-way cyclic consistency loss for editing shape or color. Yet, this method struggles to achieve precise local modification, even with a complex framework, as the color and shape are predicted in parallel, leading to potential color mismatches with the 3D shapes.
More recently, ShapeCrafter~\cite{fu2022shapecrafter} introduces an auto-regressive approach 
to progressively modify the shape 
by successively adding text. Nevertheless, they 
may not accurately
perform region-aware modifications,~\ie, unrelated regions can be significantly affected by their editing operations, as shown in Figure~\ref{fig:mani_shapecrafter}. Also, the fidelity and text consistency of both works are not satisfactory; see \new{the} results in Section~\ref{sec:experiments}.
Beyond existing works, our EXIM is 
a \new{new} hybrid explicit-implicit representation, offering explicit shape representations and shape-color disentanglement for a certain level of local modification, as well as independently modifying the color and geometry.

\section{Methodology}
\label{sec:method}

\begin{figure*}
\centering
\includegraphics[width=0.99\textwidth]{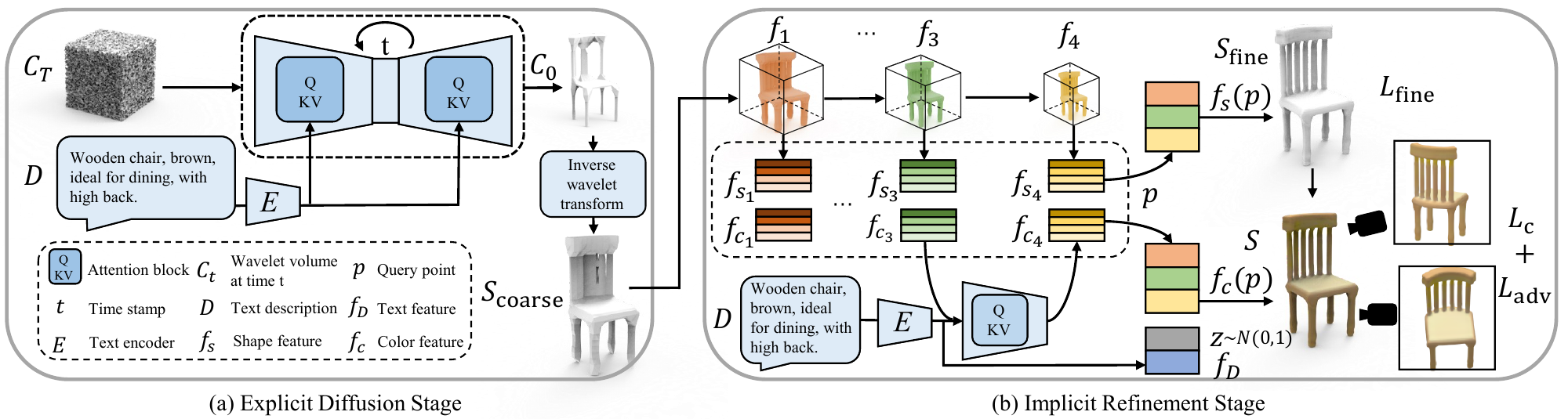}
\vspace*{-2.75mm}
\caption{Overview of our approach. (a) The Explicit Diffusion Stage (inspired by~\cite{hui2022neural}) transforms the truncated signed distance field (TSDF) of the training shape $S$ into compact wavelet volumes $C$ and utilizes a 3D diffusion model with the U-Net architecture to learn the coarse shape $S_\text{coarse}$ conditioned on the text description $D$ with cross-attention modules. (b) The Implicit Refinement Stage refines the coarse shape $S_\text{coarse}$ to produce the fine shape $S_\text{fine}$ and generates colors on the surface to create the final result $S$. It employs a 3D convolutional encoder to extract features from multi-level feature maps and predicts the occupancy and RGB values at each query location $p$. The color prediction is also conditioned on the text $D$ using cross-attention modules.
}
\label{fig:overview}
\vspace*{-2.75mm}
\end{figure*}

\subsection{Overview}
Given a text description $D$, our goal is to generate 3D shapes $S$ with colors, following $D$. 
We design a new hybrid explicit-implicit shape representation, namely EXIM, by incorporating a two-stage generation pipeline. The key in our approach is to combine the strengths of 
(i) the voxel-based explicit 3D representation for topology generation and flexible modification and 
(ii) the neural implicit representation for detail enhancement and color generation.

Figure~\ref{fig:overview} presents our framework.
In the first stage, the explicit stage, we employ  a 3D diffusion model in a compact wavelet domain to generate coarse shapes in the voxel space $S_\text{coarse}$ according to $D$. This process roughly outlines the 3D shape and enables subsequent shape modification; see Figure~\ref{fig:overview} ``Explicit Diffusion Stage''.
In the second stage,~\ie, the implicit stage, based on $S_\text{coarse}$ and text description $D$, we design an encoder network to extract multi-scale text-aware shape features. We then exploit the neural implicit function to predict the details and dress up the generated shape $S_\text{fine}$ with colors, yielding the final output $S$; see Figure~\ref{fig:overview} ``Implicit Refinement Stage''.
The two stages complement each other, enabling high-fidelity text-guided 3D shape generation. 
Further, our hybrid representation offers an explicit voxel representation and a disentangled shape and color space, allowing for a 
higher
degree of localized and independent modification of shape and color within a region of interest; see Figure~\ref{fig:overview_mani}.
Lastly, we leverage a large vision-language foundation model to extend text-to-shape generation to \new{handle} new categories \new{of shapes} without requiring human annotations, enabling a novel application of composing generated shapes for style-consistent room-level scene generation; see Figure \ref{fig:scene}. 

\subsection{Text-Guided 3D Shape Generation}\label{sec:generation}

In the following, we elaborate on our two-stage explicit-implicit model designed for text-guided 3D shape generation.

\subsubsection{Explicit Diffusion Stage} \label{sec:explicit}
In accordance with the text description $D$, the explicit stage generates coarse 3D shapes $S_\text{coarse}$ using dense voxel grids of $256^3$ by predicting the TSDF value for each voxel grid. To accomplish this, we leverage the powerful diffusion model for text-guided 3D shape generation due to its exceptional generative capabilities~\cite{ramesh2022hierarchical,vahdat2022lion}. Instead of employing the original shape space $S$, which is memory and computation-intensive, we adopt a compact wavelet space 
to perform the diffusion process\new{,} as suggested in \cite{hui2022neural}.

By utilizing biorthogonal wavelet filters~\cite{cohen1992biorthogonal}, the TSDF volume $S$, which has a size of $256^3$, is transformed into a compact, low-frequency wavelet volume $C_0$ with a size of $46^3$. By applying an inverse transform, $C_0$ can be used to reconstruct the topology of $S$ while omitting high-frequency details. Subsequently, a diffusion model is employed to generate $C_0$, which can then be converted back to $S$.
To accomplish this objective, we propose a transformer-based architecture that conditions the model on \new{the} text description $D$, \new{extending} 
the approach in~\cite{hui2022neural}. 
In the following sections, we will first provide preliminaries of diffusion models and then discuss our proposed model architecture.

Given a wavelet volume denoted as $C_0$ for a sample, the forward diffusion process incrementally adds Gaussian noise over $T$ steps, resulting in the sequence $\{C_0, C_1, ..., C_T\}$, where $C_T$ eventually becomes pure Gaussian noise following a normal distribution. Specifically, the noise data sample $C_t$ can be obtained from the following conditional probability density function:
%
\begin{equation}
q\left(\mathbf{C}_t \mid \mathbf{C}_0\right)=\mathcal{N}\left(\mathbf{C}_t ; . \sqrt{\bar{\alpha}_t} \mathbf{C}_0,\left(1-\bar{\alpha}_t\right) \mathbf{I}\right),  \bar{\alpha}_t = \prod_{i=1}^t (1-\beta_t),
\end{equation}
where $t$ \new{is} the time step and $\beta_t$ is a pre-defined noise parameter that controls the amount of noise added at step $t$; \new{$\beta_t$} increases as $t$ increases (with $\beta_T$ = 1). A diffusion model is then trained to reverse this process and recover $C_0$. 
It learns a deep denoising network $\epsilon_\theta$, which predicts the noise added at each step according to the corresponding text description $D$. Specifically, at each step $t$, The network is trained to predict the noise $\epsilon$  using \new{an} $L_2$ loss: 
\begin{equation}
\mathcal{L}_\text{coarse}=E_{t,C_0,\epsilon}[\left\| \epsilon-\epsilon_\theta(C_t,t,D \right\|^2], \epsilon\sim\mathcal{N}(0,\mathbf{I}),
\end{equation}
where $t$ \new{is} the time index, $D$ is the text description, and $\epsilon_\theta$ is a 3D U-Net designed for denoising. After training, during the inference stage, given text description $D$, we randomly sample noise $C_T$ and apply $\epsilon_\theta$ for $T$ iterations to progressively denoise the input and generate sample $C_0$, following the approach in~\cite{song2020denoising}. $C_0$ is further transformed back to the shape space to yield $S_\text{coarse}$. 

In the following, we elaborate on $\epsilon_\theta(C_t, t, D)$, which takes $C_t$, time step $t$, and a text description $D$ as inputs and produces the noise for denoising $C_t$ \new{into} $C_{t-1}$ as shown in Figure~\ref{fig:overview} (a). 
Given the text description $D$, we employ text encoder $E$ based on the CLIP-text encoder architecture~\cite{radford2021learning} to map it into word-level features $f_D \in \mathbb{R}^{L\times d}$, where $L$ is the number of words in $D$ and $d$ is the feature dimension.
Next, given $C_t \in \mathbb{R}^{46 \times 46 \times 46}$ and $f_D$ at each time step $t$, a 3D U-net with three downsampling layers and three upsampling layers is designed to produce the output. The core of the 3D U-net is a fusion module that combines the intermediate features from the U-net and the text feature $f_D$ to 
\new{encourage}
the generation of text-aligned 3D shapes.
The fusion module adopts a cross-attention mechanism~\cite{vaswani2017attention,rombach2022high} to learn the correlation between each word and each spatial location in the feature map and embeds \new{the} text features into spatial features as follows:
%
\begin{equation}
f^\prime=\text{softmax}(\frac{W_Q(f)W_K(f_D)^T}{\sqrt{d}}) W_V(f_D),
\end{equation}
where $f$ represents an intermediate feature map from the U-net with a spatial size of $H\times W$ and a feature dimension $d$. The matrices $W_Q$, $W_K$, and $W_V$ 
embed the input into query, key, and value, respectively. By doing so, the text information is locally embedded at spatial locations with high correlation, resulting in the text-aware local feature $f^\prime$ that facilitates local modification while ensuring text-shape consistency. The feature $f^\prime$ is further fed into the next layer in the U-net.
In our model, the fusion module is incorporated at the second and third downsampling layers as well as the first and second upsampling layers of our backbone architecture.

\subsubsection{Implicit Refinement Stage}

\begin{figure}
\centering
\includegraphics[width=0.99\columnwidth]{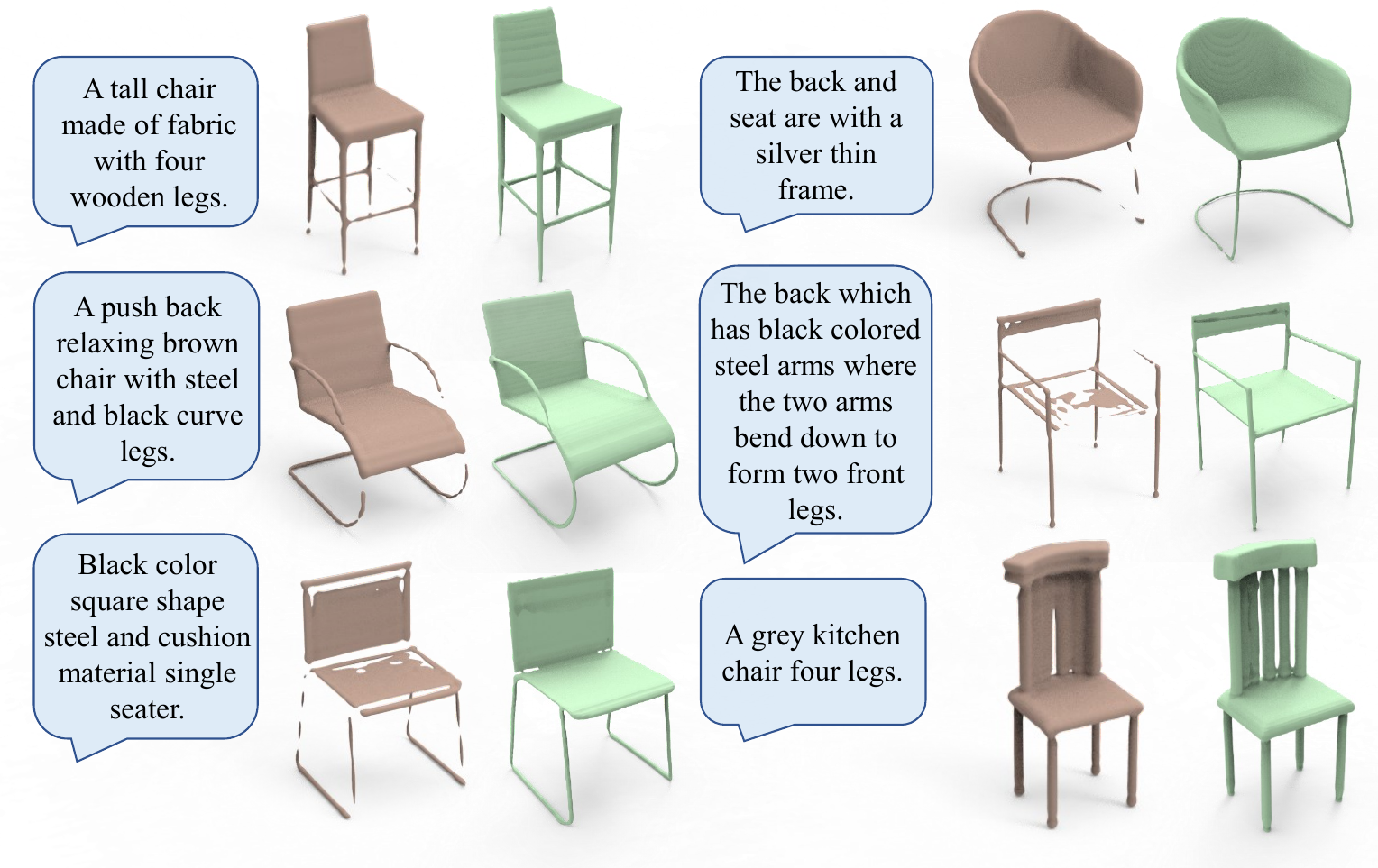}
\vspace*{-2.75mm}
\caption{The Implicit Refinement Stage (green) improves the stage-1 output (brown), particularly on the details. For \new{a} more clear comparison, color prediction is disabled.
}
\label{fig:stage1-2}
\vspace*{-2.75mm}
\end{figure}


Although the Explicit Diffusion Stage can generate plausible 3D shapes $S_\text{coarse}$, the quality of the generated details is not entirely satisfactory, as high-frequency components are discarded, limiting the model's upper-bound performance; see Figure~\ref{fig:stage1-2}. To address this issue, we propose an implicit network that compensates for the Explicit Diffusion Stage by producing high-frequency details and enabling color generation.


Specifically, given the TSDF volume of the generated 3D shape $S_\text{coarse}$ from the explicit diffusion stage, we design a 3D convolutional encoder~\cite{chibane2020implicit} for shape refinement, which extracts multi-scale hierarchical feature maps $\{f_{s_1}, \cdots, f_{s_4}\}$ from $S_\text{coarse}$ using several convolution layers and  four downsampling layers. Then, for a given query point $p_i$, we obtain its \new{associated} feature vector from different scales of feature maps and concatenate them into a feature vector $f_s(p_i) = [f_{s_1}(p_i), \cdots , f_{s_4}(p_i)]$. Following this, $f_s(p_i)$ is processed through three fully connected layers, with a sigmoid activation function at the final layer to predict the probability of the occupancy value being $1$: $o(p_i)\in (0,1)$. To train the network, we encourage the predicted occupancy values to match the ground-truth values using the sigmoid cross-entropy loss:
\begin{equation}
\mathcal{L}_\text{fine}=-\big(\sum_i \hat{o}(p_i)\log(o(p_i)) + (1 - \hat{o}(p_i))\log(1- o(p_i))\big),
\end{equation}
where $\hat{o}(p_i)$ indicates ground-truth occupancy value of $p_i$. 


For the color generation branch, we employ a 3D U-net to extract multi-scale color feature maps and incorporate the cross-attention module proposed in the explicit stage to inject word-level text features into the generated intermediate feature maps. This results in text-aware feature maps $\{f_{c_1},\cdots, f_{c_4}\}$, promoting color consistency with the text description $D$. Then, for a given query point $p_i$, we concatenate $\{f_{c_1}(p_i),\cdots,f_{c_4}(p_i)\}$, the global text feature $f_D$ from the text encoder to enhance the color and text consistency, and a random noise vector $z\sim\mathcal{N}(0,\mathbf{I})$ to encourage the generation of diverse outputs. The concatenated feature vector $f_c(p_i)$ is further fed into three fully-connected layers to produce the RGB values at query point $p_i$.
To train the color generation branch, we randomly sample a camera pose $[R|T]$ and render images $I$ and $\hat{I}$ based on the predicted colored shape and the ground-truth colored shape, respectively, using a differentiable renderer~\cite{niemeyer2020differentiable}. 
To train the model, we employ the regression loss $L_{c}$ and the adversarial loss $L_\text{adv}$ as follows: 
\begin{equation}
\mathcal{L}_c=\sum_i |I(p_i)-\hat{I}(p_i)|, \quad \text{and} \quad  \mathcal{L}_\text{adv}= \mathbb{E}_I [\log(1 - D_\phi(I))], 
\end{equation}
where $D_\phi$ represents the discriminator with parameters $\phi$. During the generative training stage, which utilizes $\mathcal{L}_c$ and $\mathcal{L}_\text{adv}$ to optimize the color generation branch, $D_\phi$ remains frozen. The discriminator is trained by maximizing $(\log D_\phi(\hat{I}) + \log(1 - D_\phi(I)))$. The adversarial loss is employed to encourage the model to produce realistic results based on the given text description $D$. 



As illustrated in Figure~\ref{fig:stage1-2}, our Implicit Refinement Stage effectively captures fine details, including long, thin structures. Moreover, this stage aids the generation of plausible colors that are consistent with both the shape $S$ and the text description $D$.

\subsection{Text-Guided 3D Shape Modification}

\begin{figure*}[h]
\centering
\includegraphics[width=0.99\textwidth]{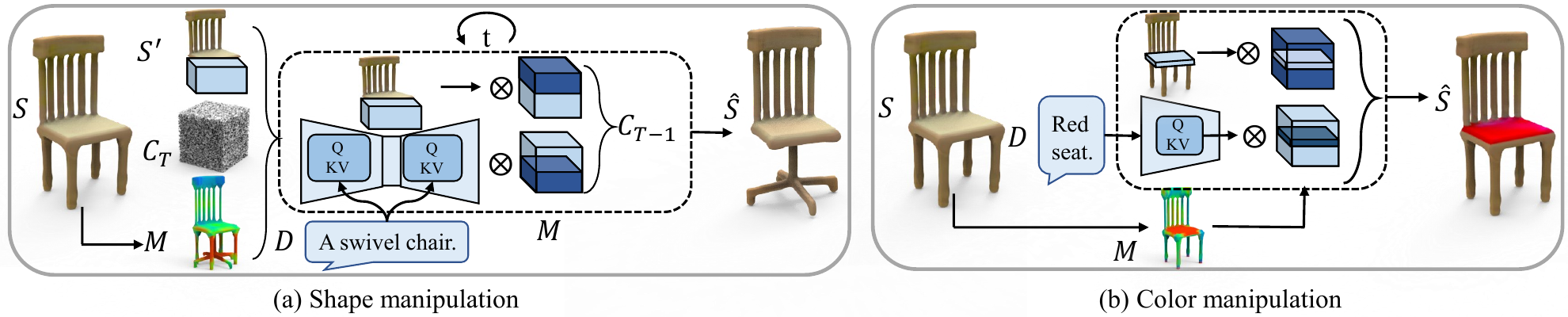}
\vspace*{-2.75mm}
\caption{
Our pipeline for modifying (a) shape and (b) color. For shape, we expand the diffusion model to include two additional input channels, the mask $M$ and the unmasked partial shape $S^\prime$. We then fine-tune the model to fill in the masked areas. For color, we can either incorporate the new text features into the intermediate feature maps or directly generate color values within the regions specified by $M$, without the need for fine-tuning.
}
\label{fig:overview_mani}
\vspace*{-2.75mm}
\end{figure*}

\begin{figure}
\centering
\includegraphics[width=0.99\columnwidth]{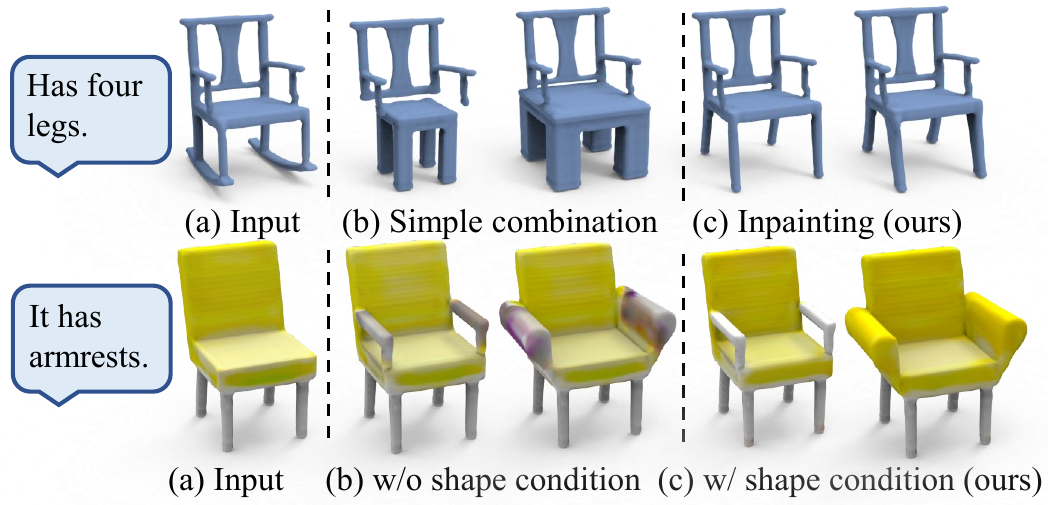}
\vspace*{-2.75mm}
\caption{
Top: directly combining the original and generated shapes may result in size and style inconsistencies, whereas our inpainting-based method effectively utilizes the context to produce consistent outcomes. Bottom: when incorporating new parts, merely copying the color from the given shape may cause noisy color predictions, while our shape-conditioned color generation generates more plausible colors.
}
\label{fig:mani_case}
\vspace*{-2.75mm}
\end{figure}


Given text description $D$, our goal is to modify input shape $S$ to align it with text $D$ while preserving \new{the} unrelated regions. Our hybrid explicit-implicit representation, \new{namely  EXIM,} provides the required flexibility for this task, as it defines a 3D topology in the explicit phase, refines the shape details, and decouples the shape and color generation. This allows us to perform a certain level of localized shape modification and also independent editing of color and shape, according to the text description $D$. 

Shape modification primarily occurs in the explicit stage, as it generates and determines the shape's topology explicitly, allowing for the injection of certain local shape information,
It is important to note that the wavelet volume corresponds locally to the original shape, meaning that modifying the data in wavelet domain is equivalent to that modifying the data in the original domain. Figure~\ref{fig:overview_mani} 
illustrates how we may modify the shape and color.
First, given 3D shape $S$ and text description $D$, we generate mask $M$ that localizes the regions of interest to be modified. To create mask $M$, one can either manually select/refine an existing mask or generate it using our pre-trained explicit diffusion model in Section~\ref{sec:generation}, following DiffEdit~\cite{couairon2022diffedit}; details are given in the supplementary material.
Second, with mask $M$, a naive approach for modifying the shape would be to directly generate a new shape $\bar{S}$ using text $D$ and combine it with the existing shape $S$ based on $M$. However, size and style may not be consistent, as seen in Figure~\ref{fig:mani_case} top (b).

To this end, we adapt our pre-trained explicit diffusion model for text-guided shape inpainting by augmenting the input wavelet volume $C_t$ with two additional channels: mask $M$ and shape $S^\prime$, which represents the original shape $S$ with the masked areas excluded. We initialize the two additional channel weights to zero and use the pre-trained network weights to initialize the remaining parameters. Then, we fine-tune the model to paint the masked areas and produce coherent manipulated shapes according to text $D$.
During the fine-tuning, we simulate mask $M$ using the unsupervised part-segmentation approach~\cite{chen2019bae} to produce part-level masks and train the model to paint the masked regions based on the original text $D$ to emulate the scenarios at inference. After the training, the model can enable a certain level of localized shape modification, according to the text and generate coherent shapes, as demonstrated in Figure~\ref{fig:mani_case} top (c).

 The edited shape $\hat{S}$ is further processed by the Implicit Refinement Stage for detail refinement and color generation. Notably, our implicit refinement stage is conditioned on shape $\hat{S}$, enabling the production of colors consistent with the edited 3D shape. As depicted in Figure~\ref{fig:mani_case} bottom (c), our method creates plausible colors that align with the manipulated shape, while other approaches fail to produce well-aligned colors, as seen in Figure~\ref{fig:mani_case} bottom (b).
 


Furthermore, our decoupled design for coarse shape and color generation enables us to independently manipulate color without affecting the shape. Given mask $M$ and text description $D$ for color editing, we integrate the new text features into the intermediate feature maps within the regions determined by $M$, producing colors in accordance with \new{text} $D$. This allows us to manipulate color within the regions of interest without impacting the shape or other areas, as demonstrated in Figure \ref{fig:overview_mani} ``color modification''.

\subsection{Applications on New Categories and Indoor Scene Generation}

Our model requires paired text-shape data~\cite{chen2018text2shape} for training, which is challenging to obtain through human annotations. Therefore, we \new{further} explore the possibility of expanding existing text-to-shape annotations to new categories by leveraging pre-trained vision-language foundation models, such as BLIP-v2~\cite{li2023blip}. Specifically, we render an image for a given shape and employ a foundation model (\ie, BLIP-v2~\cite{li2023blip}) for image captioning to generate a pseudo text description of the shape. Despite its simplicity, this strategy can effectively generate descriptions such as ``\textit{The chair is a modern, two-seater sofa with a metal frame and a red fabric seat}''. 
More analysis of the quality of \new{the} pseudo annotation\new{s} is \new{provided} in the supplementary material. 
This approach allows us to expand the text-to-shape dataset~\cite{chen2018text2shape} with new categories, further enabling novel applications in scene-level generation, as described below.

We investigate the potential of using text-to-shape generation to create style-consistent room-level scenes by composing generated 3D shapes. Given the generated shapes for room assets, such as tables, chairs, sofas, and cabinets, we employ a 2D room layout map, which can be obtained using existing methods~\cite{paschalidou2021atiss} or human design, to determine the object location, orientation, and size. Subsequently, the shapes are arranged accordingly to generate a room. It is crucial to note that we can promote the production of style-consistent shapes by using text prompts with similar descriptions for desired color/shape styles. 
Our preliminary attempt demonstrates that our method can effectively facilitate the creation of indoor scenes with high consistency, as shown in Figure~\ref{fig:scene}.

\begin{figure*}[h]
\centering
\includegraphics[width=0.99\textwidth]{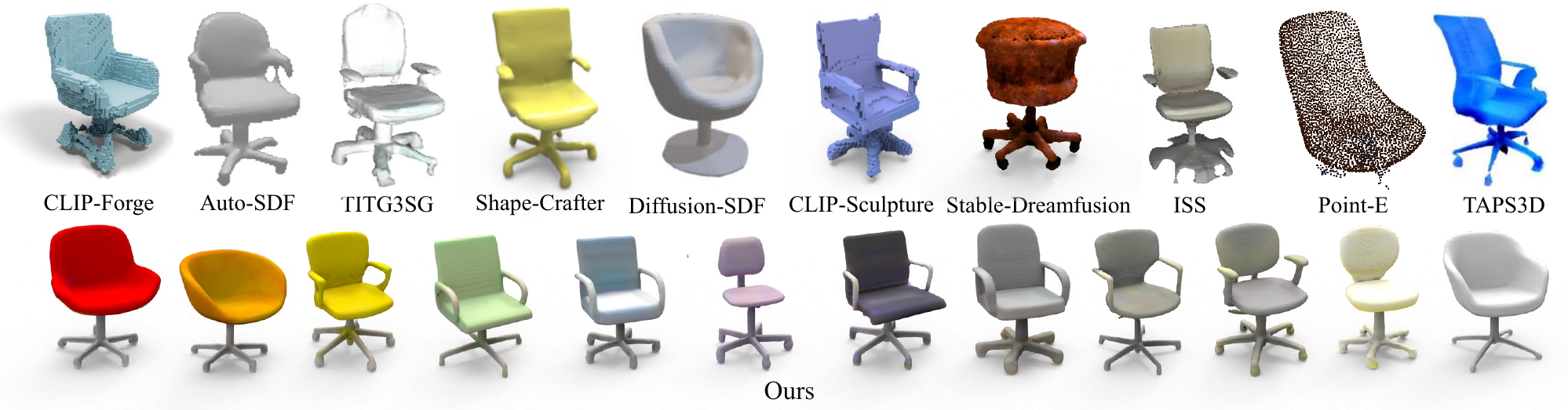}
\vspace*{-2.75mm}
\caption{ Swivel chairs generated by various approaches. Top row: results from existing works. The outcomes of Stable-Dreamfusion~\cite{stable-dreamfusion} and Point-E~\cite{nichol2022point} are obtained using their official codes, while the results from other methods are directly taken from their respective papers. Bottom row: our method. High resolution figures are provided in the supplementary material.  
}
\label{fig:swivel}
\vspace*{-2.75mm}
\end{figure*}

\begin{figure*}[h]
\centering
\includegraphics[width=0.99\textwidth]{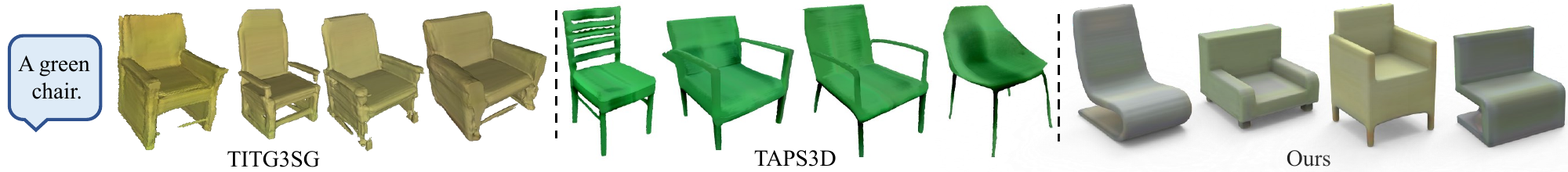}
\vspace*{-2.75mm}
\caption{Comparison with TITG3SG~\cite{liu2022towards} and TAPS3D~\cite{wei2023taps3d}. High resolution figures are provided in the supplementary material. }
\label{fig:diversity}
\vspace*{-2.75mm}
\end{figure*}

\section{Experiments}
\label{sec:experiments}

\subsection{Dataset and Implementation Details}

We train our framework on five ShapeNet categories~\cite{shapenet2015}, including chair, table, sofa, cabinet, and airplane. For the chair and table categories, we train one model with pseudo annotations and another with annotated texts provided by~\cite{chen2018text2shape} to show the efficacy of the generated pseudo annotations. For the remaining categories, we train models using pseudo annotations.
Prior to model training, we convert the mesh to TSDF and subsequently to wavelet volumes, following~\cite{hui2022neural}. Initially, we train the Explicit Diffusion Stage for 2,000 epochs using the Adam optimizer and then train the Implicit Optimization Stage \new{additionally for} 200 epochs with an initial learning rate of $1e^{-4}$.
For shape modification, we expand the input to three channels and fine-tune the model for 500 epochs using the same learning rate. Following~\cite{hertz2022spaghetti, luo2021diffusion, hui2022neural}, we train a separate model for each category.
We employ classifier-free diffusion guidance~\cite{ho2022classifier} in both the Explicit Diffusion Stage and shape modification fine-tuning. Empirically, we set the guidance weight $g=3$.
To demonstrate the efficacy of our approach, we primarily conduct experiments on the chair category due to its high complexity and diverse topologies and colors.


\subsection{Comparisons with Existing Works}

\paragraph{Text-guided 3D \new{S}hape Generation}

To evaluate the performance of our \new{EXIM}, we compare it with existing works both qualitatively and quantitatively. As illustrated in Figure~\ref{fig:swivel}, our approach significantly outperforms all \new{the} existing works~\cite{sanghi2021clip, mittal2022autosdf, liu2022towards, fu2022shapecrafter, li2022diffusion, sanghi2022textcraft, poole2022dreamfusion, liu2023iss, nichol2022point, wei2023taps3d}. 
Specifically, CLIP-Forge, Auto-SDF, Diffusion-SDF, CLIP-Sculpture, ISS, and Point-E cannot generate high-quality structured details,~\eg, wheels. TITG3SG and ShapeCrafter produce asymmetrical arms. Stable-Dreamfusion produces an unrealistic seat and requires long-time optimization, \new{whereas} TAPS3D creates unnatural textures. 
\new{Also}, none of the above existing works can create high-quality and smooth surfaces. On the contrary, our model is able to generate diverse, high-quality 3D shapes in terms of structured details, colors, and surfaces. 
Moreover, Figure~\ref{fig:diversity} demonstrates our model's diversified generative capability from a single given text description, surpassing~\cite{liu2022towards, wei2023taps3d} in terms of both shape and color.


To quantitatively compare our \new{EXIM} with \new{the} existing works, we adopt CLIP-S\new{,} following~\cite{fu2022shapecrafter}\new{,} to measure \new{the} text-shape consistency and use FPD and FID~\cite{wei2023taps3d, heusel2017gans} to evaluate the generative quality of shape and color, respectively. For a fair comparison, we adopt the results of existing works reported in ~\cite{fu2022shapecrafter} and~\cite{wei2023taps3d}. Table~\ref{tab:quanComparison} \new{shows} that our model consistently outperforms \new{the} existing works across all metrics, indicating \new{its} superior performance 
in terms of generative quality and text consistency. ``N.A.'' means the result is not reported in~\cite{fu2022shapecrafter} and~\cite{wei2023taps3d}.
\

\begin{table}[t]
	\centering
		\caption{Comparison with existing works. 
  }
        \vspace*{-2mm}
	\resizebox{\linewidth}{!}{
		\begin{tabular}{C{4.1cm}|C{1.4cm}|C{1.0cm}|C{1.0cm}}
			\toprule[1pt]
                        Method &   CLIP-S ($\uparrow$)   & FPD  ($\downarrow$) & FID ($\downarrow$)  \\ \hline
                         Text2Shape~\cite{chen2018text2shape} & 16.27& N.A.& N.A.  \\
                         CLIP-Nerf~\cite{wang2021clipnerf} & N.A. & N.A. & 48.4 \\
                         CLIP-Forge~\cite{sanghi2021clip} & 26.34 & 825.96 & N.A. \\
                         TITG3SG~\cite{liu2022towards} & 38.88 & 1566.76 & N.A. \\
                         AutoSDF~\cite{mittal2022autosdf} & 48.92& N.A.& N.A. \\
                         ShapeCrafter~\cite{fu2022shapecrafter} & 52.43&N.A.& N.A. \\
                         TAPS3D~\cite{wei2023taps3d} & N.A. & 342.23 &  43.7 \\
\hline
EXIM (Ours)  & \textbf{71.75}  & \textbf{192.17} &  \textbf{36.8} \\
			\bottomrule[1pt]
	\end{tabular}}
    \vspace*{-1mm}
\label{tab:quanComparison}
\end{table}

\paragraph{Text-guided 3D Shape Modification}

\begin{figure*}[h]
\centering
\includegraphics[width=0.99\textwidth]{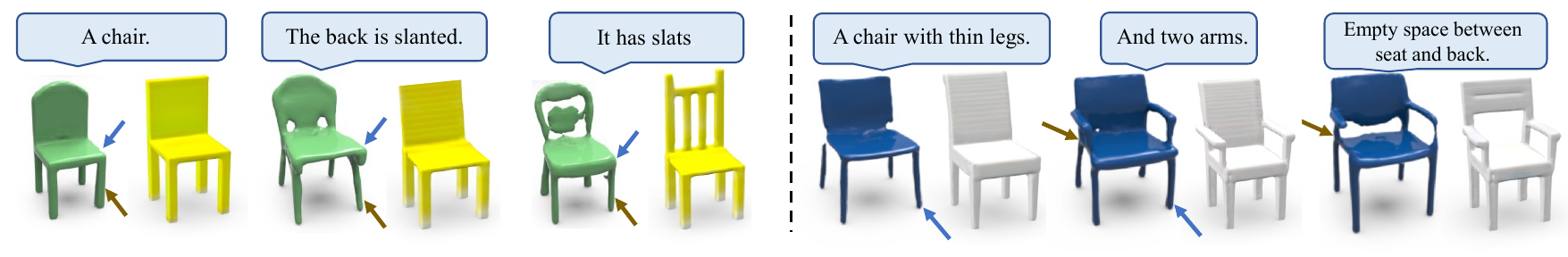}
\caption{
Comparison with ShapeCrafter~\cite{fu2022shapecrafter} for text-guided modification. Green and blue: ShapeCrafter. Yellow and white: Ours. Left: ShapeCrafter alters the legs (red arrow) and the seat (blue) while editing the back. Right: ShapeCrafter modifies the legs (blue) when adding armrests and adjusts one armrest (red) while editing the back. Our approach maintains unmentioned regions unaltered. Note that our color prediction is disabled for \new{a} fair comparison. 
}
\label{fig:mani_shapecrafter}
\vspace*{-2.75mm}
\end{figure*}

\begin{figure*}
\centering
\includegraphics[width=0.99\textwidth]{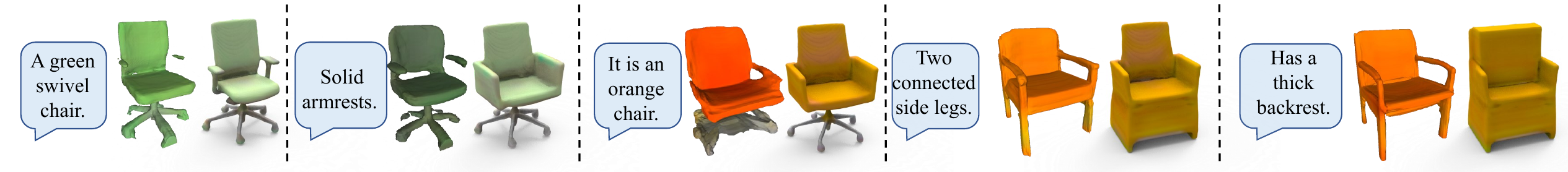}
\caption{Text-guided modification} using TITG3SG~\cite{liu2022towards} (left) and  our method (right).
\label{fig:mani_cvpr22}
\vspace*{-2.75mm}
\end{figure*}

Next, we compare the text-guided modification capabilities of our EXIM with existing works  ShapeCrafter~\cite{fu2022shapecrafter} and TITG3SG~\cite{liu2022towards}. As Figures~\ref{fig:mani_shapecrafter} and~\ref{fig:mani_cvpr22} \new{show}, our approach \new{allows more accurate editing of} the regions of interest according to the text descriptions, whereas the two baselines may significantly impact other regions \new{when they modify the shapes.}
For example, ShapeCrafter alters the seat and legs when editing the back; see Figure~\ref{fig:mani_shapecrafter} top row, \new{whereas} TITG3SG cannot keep the original color when editing the shape and may alter the shape when \new{modifying the} color.

More importantly, our \new{EXIM} is able to modify both the geometry and color of 3D shapes using text descriptions successively, as Figure~\ref{fig:manipulation} \new{shows}. For instance, we can easily convert the legs of a chair to wheels and add slats in the backrest of the chair without affecting the other unmentioned regions.  
More 
results are \new{provided} in the supplementary material.



\begin{figure*}
\centering
\includegraphics[width=0.99\textwidth]{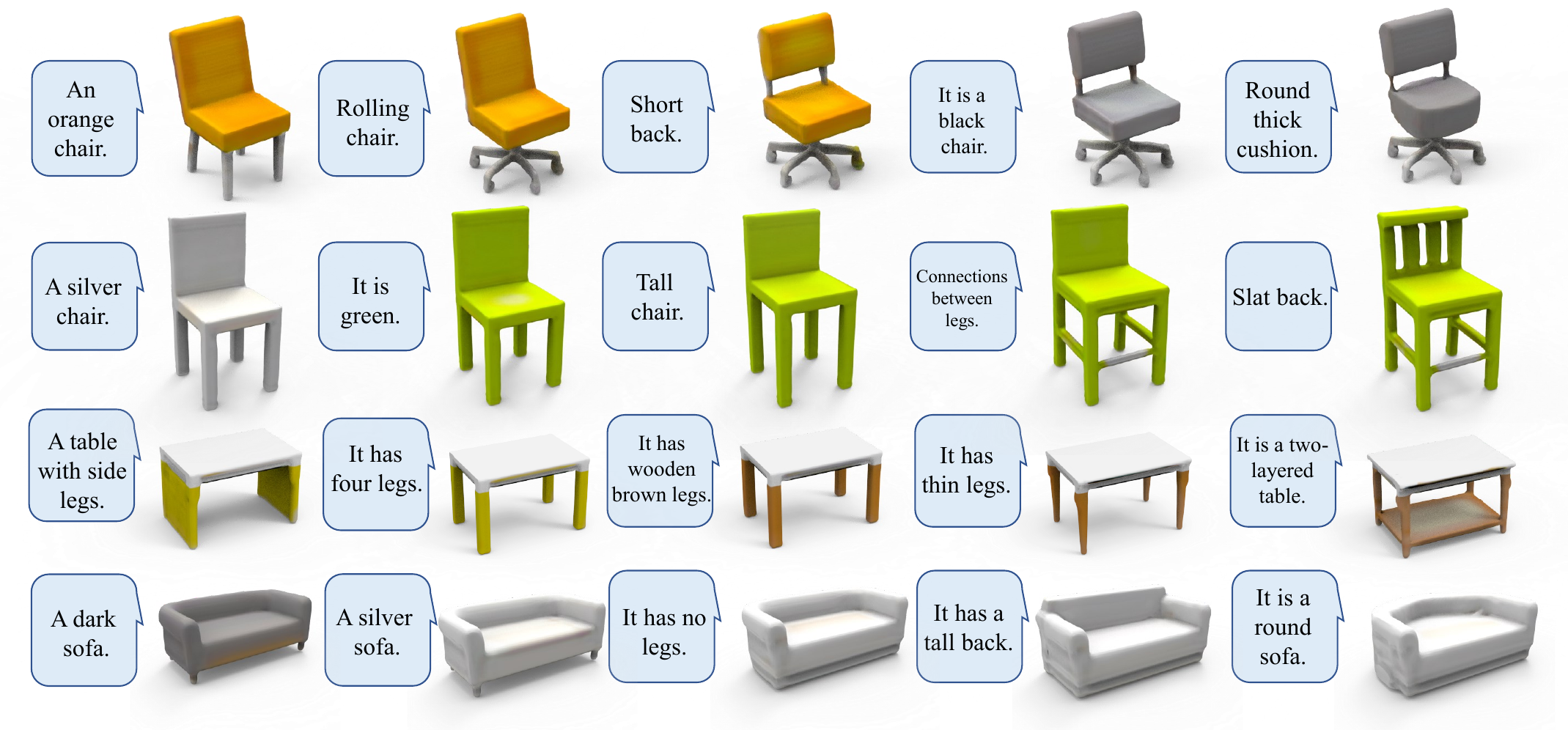}
\caption{Our approach enables us to successively
modify 3D shapes using the text descriptions.}
\label{fig:manipulation}
\vspace*{-2.75mm}
\end{figure*}

\subsection{Indoor Scene Generation}
\begin{figure*}
\centering
\includegraphics[width=0.99\textwidth]{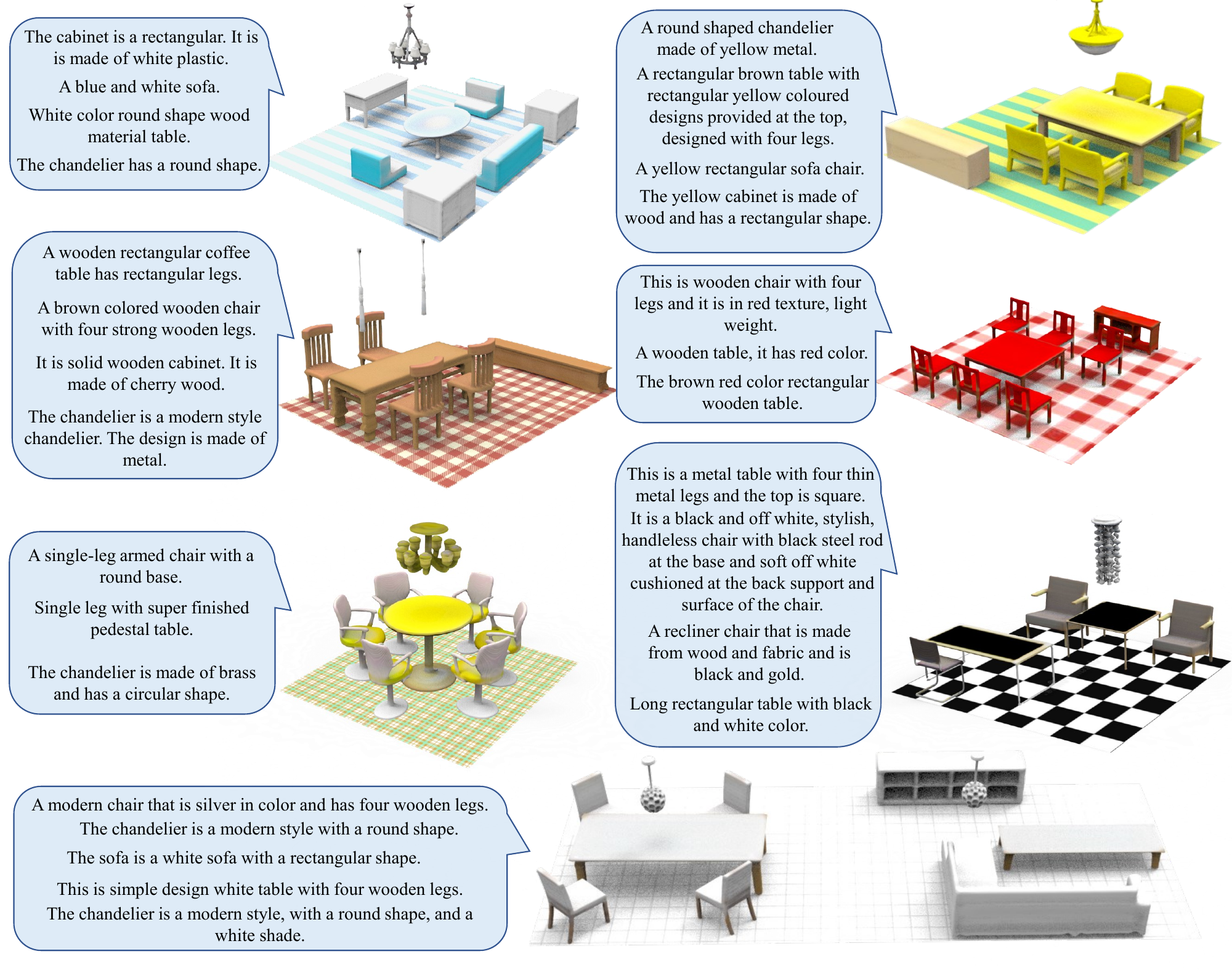}
\caption{Our approach effectively generates style-consistent indoor scenes by utilizing \new{the} text descriptions, considering both \new{the} color and shape aspects. High resolution figures are provided in the supplementary material. 
}
\label{fig:scene}
\vspace*{-2.75mm}
\end{figure*}

We extend our approach to indoor scene generation by composing shapes generated 
by conditioning the process
on text prompts. 
As illustrated in Figure~\ref{fig:scene}, our approach can generate style-consistent room-level 3D scenes in terms of both color and shape.


\subsection{Ablation Studies and Generation Novelties}

\paragraph{Analysis on Word-level Cross-Attention Modules}

\begin{figure*}
\centering
\includegraphics[width=0.99\textwidth]{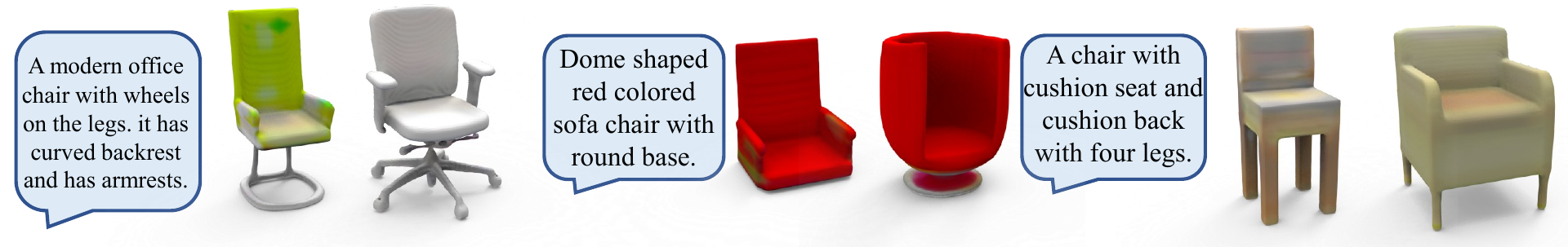}
\caption{Generative results with feeding a text feature to the network (left) vs. our cross-attention module (right). }
\label{fig:ablation0}
\vspace*{-2.75mm}
\end{figure*}

To validate the effectiveness of the word-level cross-attention module in embedding localized information into features, we design a baseline that directly feeds the global feature of the text description $f_D$ into the network without using the attention module. 
Specifically, we transform $f_D$ to the same feature dimension as the feature map $F_i$ using an MLP and multiply $F_i$ with $f_D$. Figure~\ref{fig:ablation0} shows that our approach\new{,} with word-level cross-attention\new{,} can produce results \new{that are more consistent} with the text descriptions. 


\paragraph{Analysis on the Implicit Stage}

\begin{figure*}
\centering
\includegraphics[width=0.99\textwidth]{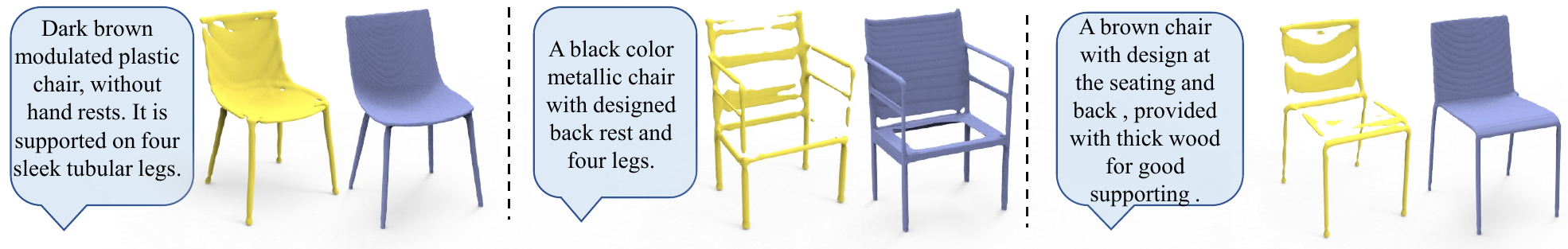}
\vspace*{-2.75mm}
\caption{\new{The} explicit stage (violet) refines the output of the Implicit stage (yellow). Color prediction is disabled for more clear comparison\new{s}.}
\label{fig:ablation1}
\vspace*{-1.75mm}
\end{figure*}

To investigate the manner in which the implicit stage refines the shape, we conduct a comparative analysis of the generative quality produced by the two stages \new{of EXIM}. As illustrated in Figure~\ref{fig:ablation1}, the implicit refinement stage effectively recovers high-frequency details such as thin components, which are challenging to capture using the hand-crafted wavelet components $C_0$. To quantitatively assess the generative details, we employ \new{the} $COV$ and $MMD$ metrics, as they can reveal the quality of these details. The findings presented in Table~\ref{tab:ablation1} demonstrate that the results refined by the implicit stage consistently surpass 
the stage-1 output\new{s}.


\paragraph{Analysis on Multi-scale Feature Fusion}

\begin{figure*}
\centering
\includegraphics[width=0.99\textwidth]{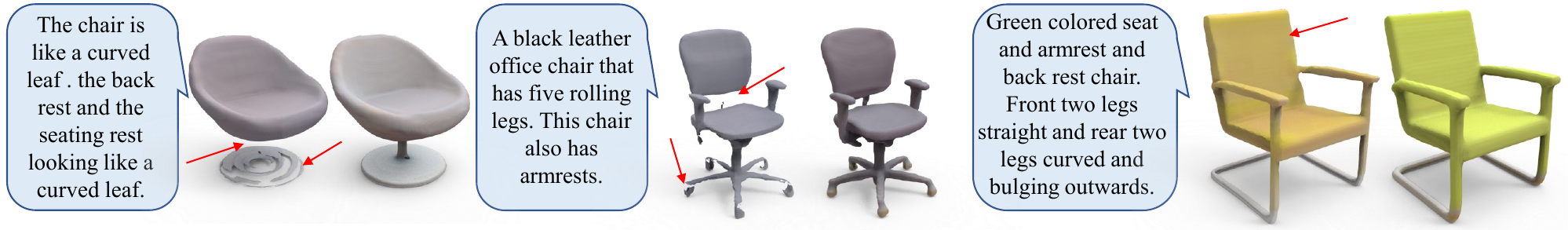}
\vspace*{-2.75mm}
\caption{Visual ablation on the generated results from a single scale feature (left) vs. ours multi-scale features (right) in the Implicit Stage. The single-scale feature may lead to artifacts, such as those indicated by the red arrows. Left: missing leg and noisy bottom. Middle: missing connections and incomplete wheels. Right: the yellow color is inconsistent with the words ``green colored'' in the input text. }
\label{fig:single_scale}
\vspace*{-1.75mm}
\end{figure*}

To assess the effectiveness of our multi-scale feature fusion approach within the Explicit Stage, we establish a baseline model. This baseline model predicts the occupancy and RGB values using the preceding feature map exclusively, rather than merging all the previous features from $f_1$ to $f_4$. As illustrated in Figure~\ref{fig:single_scale}, our multi-level feature fusion strategy blends the semantic-rich deep features with detail-rich shallow features. This integration helps to generate detail-preserved structures and colors that are more consistent with the input texts.

\paragraph{Analysis on Modification} Designs

\begin{figure*}
\centering
\includegraphics[width=0.99\textwidth]{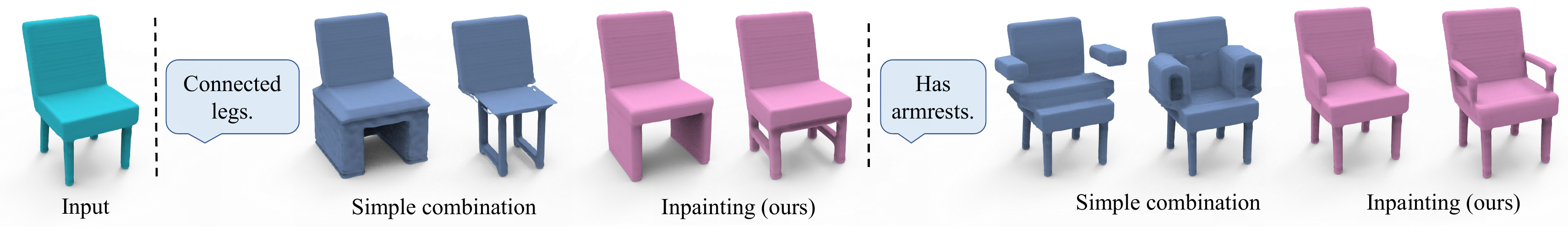}
\vspace*{-2.75mm}
\caption{
In comparison to the \new{simple} 
combination using the mask (blue), our shape inpainting method (pink) enhances the consistency between the given and generated shapes. Colors are used for visualization purposes. 
}
\label{fig:ablation2-1}
\vspace*{-2.75mm}
\end{figure*}

\begin{figure*}
\centering
\includegraphics[width=0.99\textwidth]{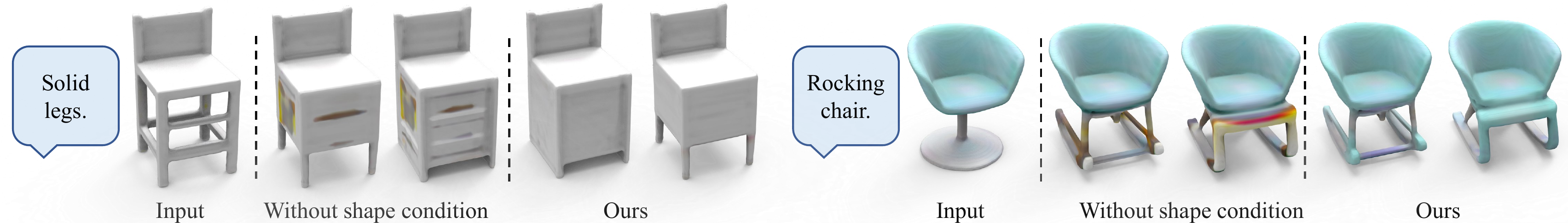}
\caption{Our color prediction is conditioned on the predicted shape, enhancing the shape-color consistency \new{when modifying the shape}.}
\label{fig:ablation2-2}
\vspace*{-2.75mm}
\end{figure*}

First, we examine the efficacy of our inpainting strategy for shape modification. As depicted in Figure~\ref{fig:ablation2-1}, our approach, in comparison to directly combining the generated results with the given shape using mask $M$, is capable of generating consistent modification results in terms of both size and style, for instance, in the case of legs and arms.
Next, we explore the impact of the shape-color decoupled generation strategy on modification. As illustrated in Figure~\ref{fig:ablation2-2}, the newly created structures resulting from shape modification exhibit noisy color due to a lack of shape awareness. \new{Yet}, our approach \new{is able to} effectively produce plausible colors more consistently with the manipulated shapes.


\begin{table}[t]
	\centering
		\caption{Ablation study on \new{the} hybrid representation. The generative quality of our Implicit Stage consistently outperforms
  the Explicit Stage, which is inspired by~\cite{hui2022neural}}. 
        \vspace*{-2mm}
	\resizebox{1\linewidth}{!}{
		\begin{tabular}{C{3cm}|C{1cm}C{1cm}C{1cm}C{1cm}}
			\toprule[1pt]
                        \multirow{2}*{Method} & 
                          \multicolumn{2}{c}{COV ($\uparrow$)} & \multicolumn{2}{c}{MMD ($\downarrow$)} \\
                         & CD         & EMD        & CD         & EMD  \\ \hline
Explicit Stage
& 39.8          & 38.7 & 14.0          & 16.3    \\ \hline 
Implicit Stage & 
\textbf{41.7} & \textbf{42.4} & \textbf{13.8} & \textbf{15.9} \\ \hline 
			\bottomrule[1pt]
	\end{tabular}}
    \vspace*{-1mm}
\label{tab:ablation1}
\end{table}

\begin{figure*}
\centering
\includegraphics[width=0.99\textwidth]{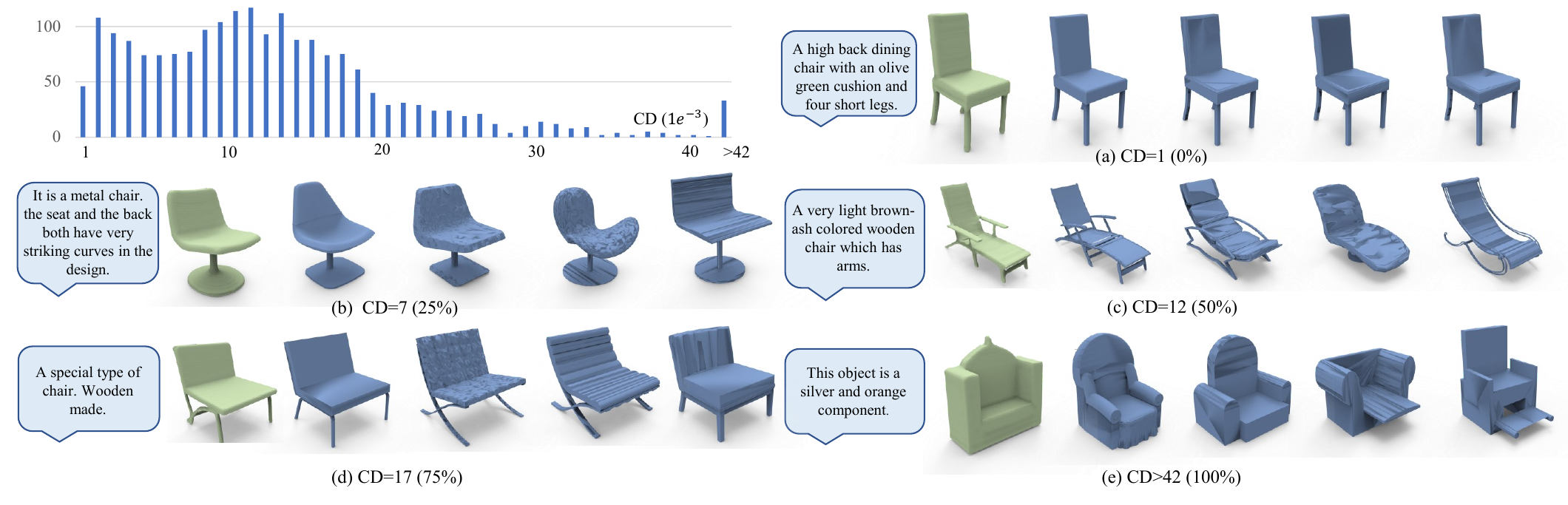}
\caption{
Analysis of generation novelty. For each of the 2,000 generated chairs, we identify the most similar chairs in the training data with the smallest Chamfer Distance (CD). Top left: the distribution of CD values for all 2,000 generated chairs in relation to their nearest neighbors in the training data. Others: for each generated chair (green), we retrieve the top four most similar 
\new{chairs} (blue). Notably, significant differences exist between our generated \new{chairs}
and the retrieved ones for 75\% (from (b) to (e)) of the generated samples with $CD\geq7$. Color prediction is disabled \new{for more clear comparisons on the shapes}. 
}
\label{fig:novelty}
\vspace*{-2.75mm}
\end{figure*}

\paragraph{Generation Novelties} 
We analyze 
our method's ability to generate novel shapes instead of merely replicating samples from the training set. As illustrated in Figure~\ref{fig:novelty}, we plot the distribution of Chamfer Distance (CD) between each generated shape and its nearest neighbors within the training set. In Figure~\ref{fig:novelty}, for each given generated sample (the ``green'' shape), we display its four nearest neighbors from the training samples, arranged from the most similar to the least similar, from left to right. It is important to note that a substantial difference between the generated and retrieved shapes exists at $CD=7 (25\%)$. Here, $CD=7 (25\%)$ means  that only 25\% of all evaluated samples have a Chamfer distance to their nearest neighbors smaller than $7$. This finding indicates that a significant portion of the generated results from our approach comprises novel shapes that go beyond the training set.

\subsection{Additional Generative Results}

\begin{figure*}
\centering
\includegraphics[width=0.99\textwidth]{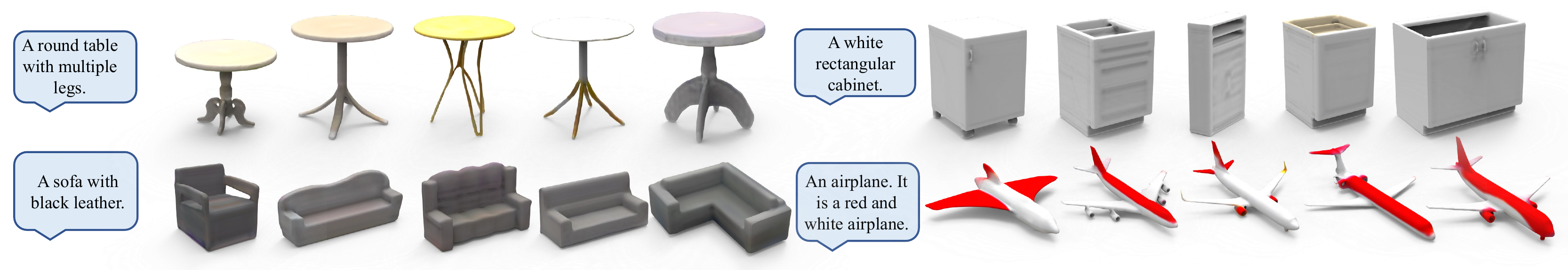}
\caption{Diversified generative results of our approach. High resolution figures are provided in the supplementary material. }
\label{fig:results}
\vspace*{-2.75mm}
\end{figure*}

We present additional generative results for multiple categories using our pseudo annotations in Figure~\ref{fig:results}. The pseudo text enables us to generate a diverse array of 3D shapes that are consistent with input texts, eliminating the need for human-annotated texts. More generative results are illustrated in the supplementary material.

\vspace{-1mm}

\section{Conclusion}
\label{sec:conclusion}

We presented a new text-guided 3D shape generation approach. The key contribution of this work is a hybrid explicit-implicit (EXIM) 3D shape representation that combines the strengths of explicit and implicit shape representations. Specifically, the explicit stage learns the topology of 3D shapes and enables a certain level of local modification, while the implicit stage adds the details and predicts the colors. Additionally, the shape and color predictions are decoupled and color prediction is conditioned on shape to promote shape-color consistency in the modification.
Our approach surpasses existing works in its ability to create high-quality 3D shapes and to facilitate local modifications using texts. Also, we extend our approach to additional categories by leveraging image captioning models to enable the innovative application of generating style-consistent room-level scenes.
Extensive experiments demonstrate that our approach outperforms existing works regarding generative quality and text-shape consistency. We hope our investigation will inspire further research in text-to-shape generation and dataset creation. Limitations are discussed in the supplementary material.

\begin{acks}
The work has been supported by the Research Grants Council of the Hong Kong Special Administrative Region (Project No. CUHK 14206320,14201921) and the InnoHK of the Government of the Hong Kong Special Administrative Region via the Hong Kong Centre for Logistics Robotics. 
\end{acks}

\vspace{+20mm}
\centerline{\Large{\textbf{Supplementary Material}}}

\section{Additional Generation Results}

\begin{figure*}
\centering
\includegraphics[width=0.99\textwidth]{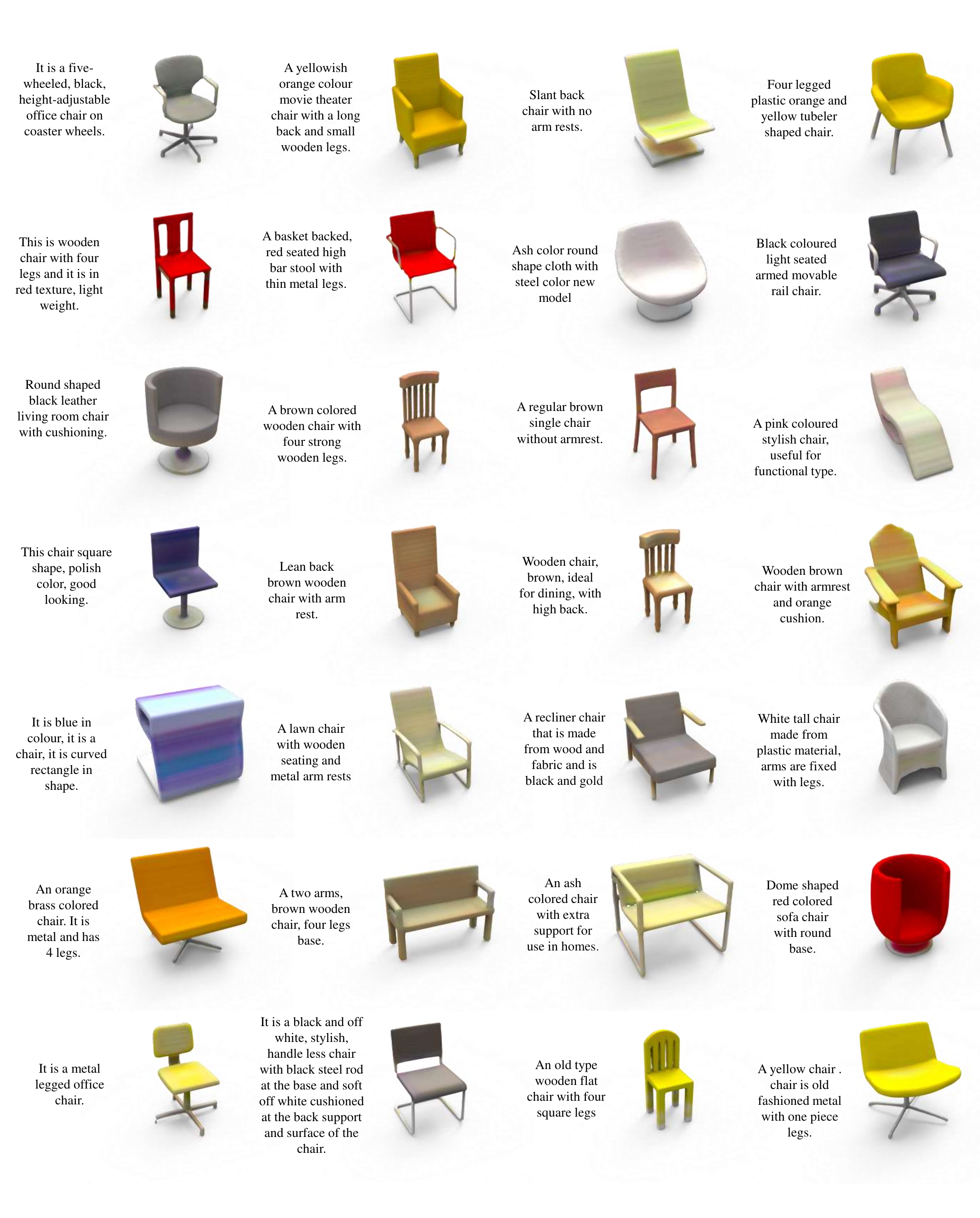}
\caption{Additional text-guided generation results on the chair category. 
}
\label{fig:chair}
\vspace*{-2.75mm}
\end{figure*}

\begin{figure*}
\centering
\includegraphics[width=0.99\textwidth]{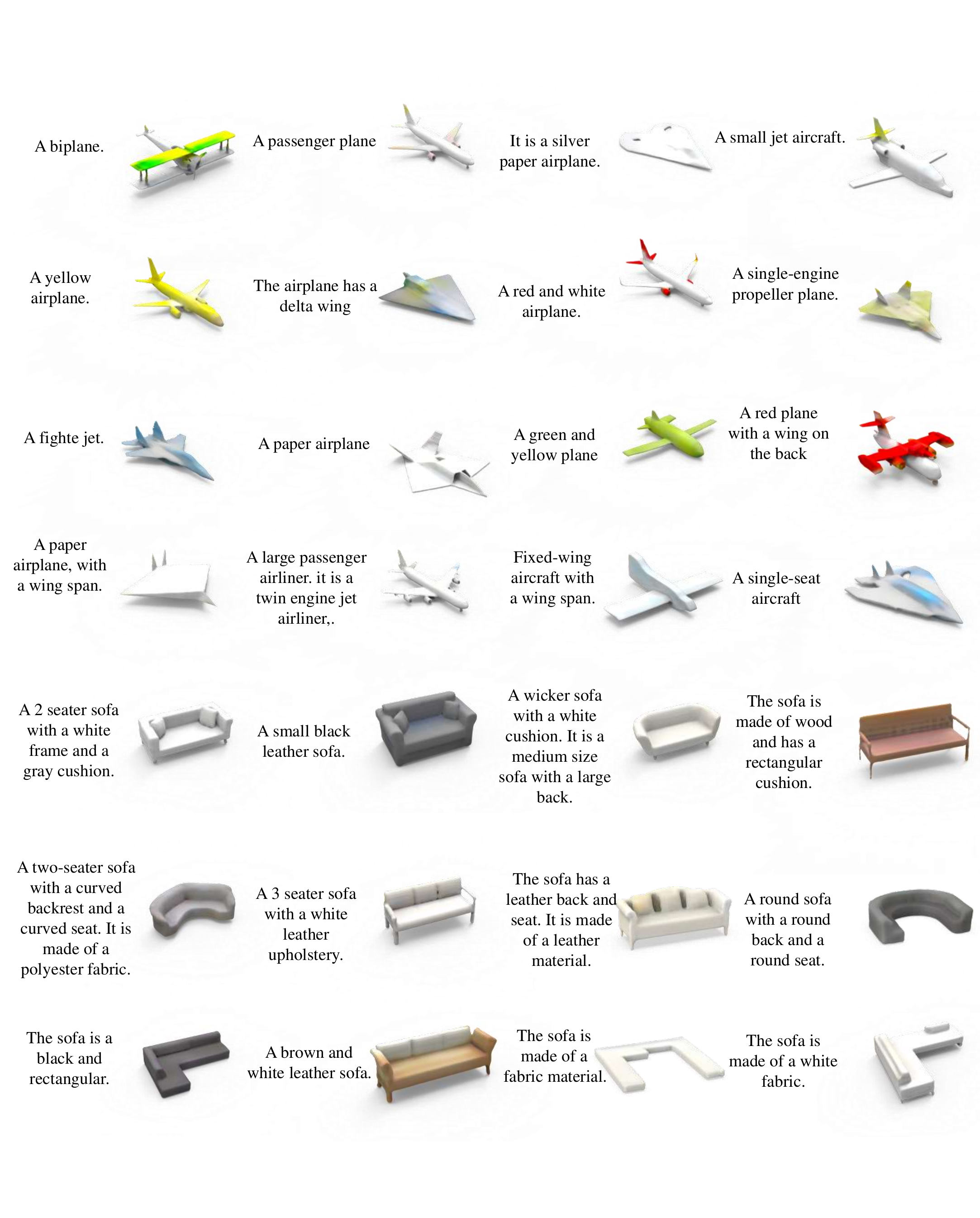}
\caption{Additional text-guided generation results on the airplane and sofa categories. 
}
\label{fig:air}
\vspace*{-2.75mm}
\end{figure*}

\begin{figure*}
\centering
\includegraphics[width=0.99\textwidth]{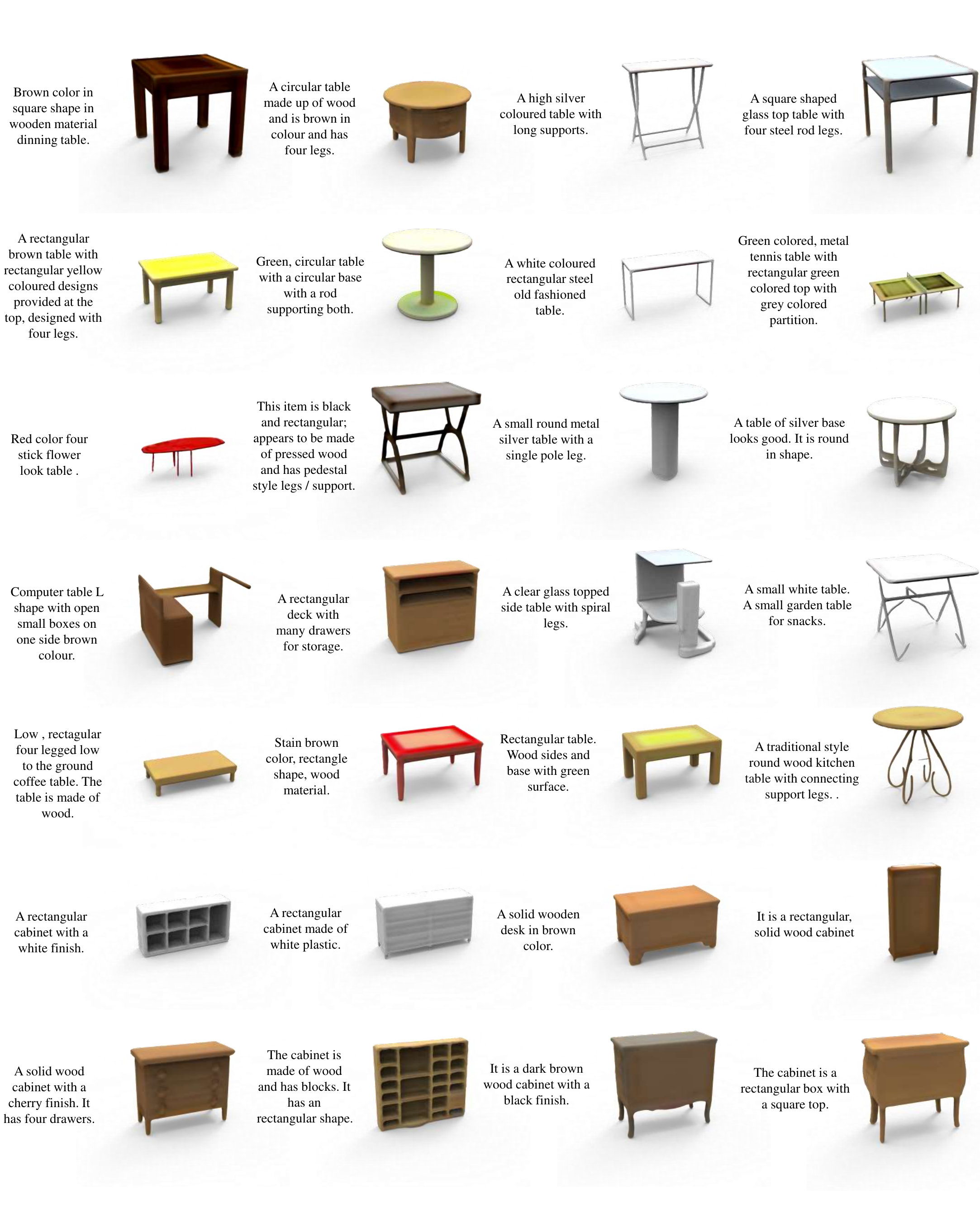}
\caption{Additional text-guided generation results on the table and cabinet categories. 
}
\label{fig:table}
\vspace*{-2.75mm}
\end{figure*}

\begin{figure}
\centering
\includegraphics[width=0.99\columnwidth]{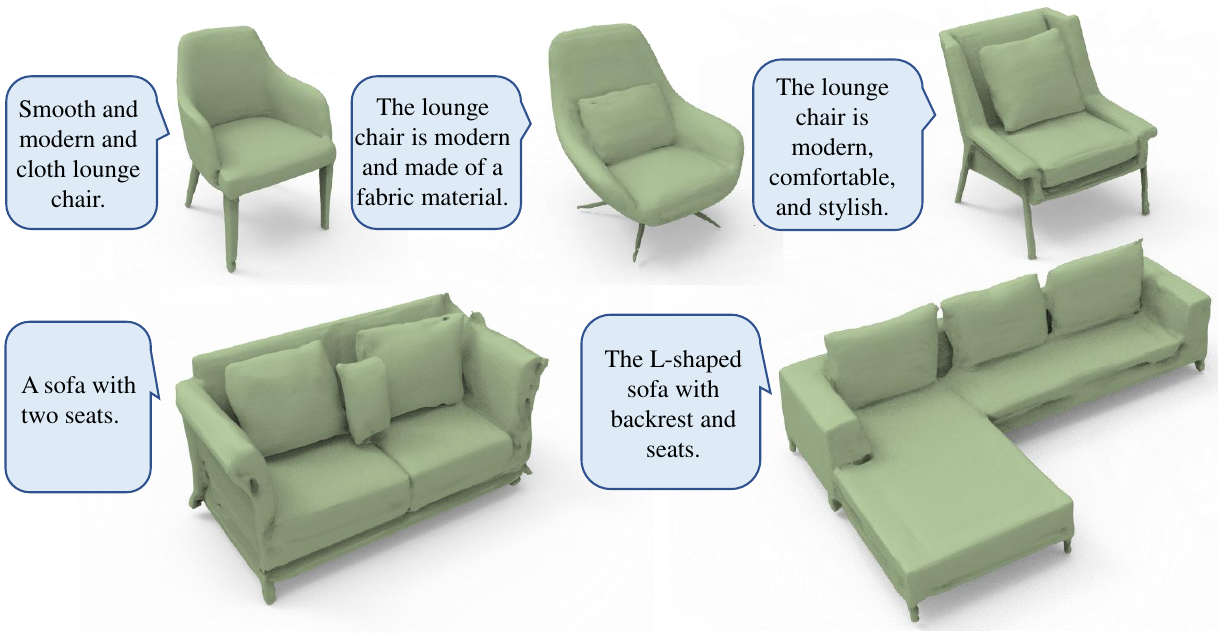}
\caption{Results produced by our model trained on the 3D-FUTURE dataset~\cite{fu20213d}. 
}
\label{fig:3dfuture}
\vspace*{-2.75mm}
\end{figure}

First, we show more generations results from texts in Figures~\ref{fig:chair}, Figure~\ref{fig:air}, and Figure~\ref{fig:table}. 
When trained on the 3D-FUTURE dataset~\cite{fu20213d}, our approach can generate more detail-rich shapes; see pillows and cushions in the chairs and sofas of Figure~\ref{fig:3dfuture}. 

These results further demonstrate our EXIM can effectively create high-fidelity 3D shapes from texts that are highly consistent with the input texts. 

\section{Additional Moification Results}

We further show additional text-guided 3D shape modification results. As shown in Figure~\ref{fig:manipulation}, we can effectively edit a local region (swivel chair, Z-shape leg, S-shape leg, and X-cross legs) and topology (tall legs and short legs) using texts without affecting other unrelated regions. Note that we show more results of shape modification since it is a more challenging task than editing colors.

\begin{figure*}
\centering
\includegraphics[width=0.99\textwidth]{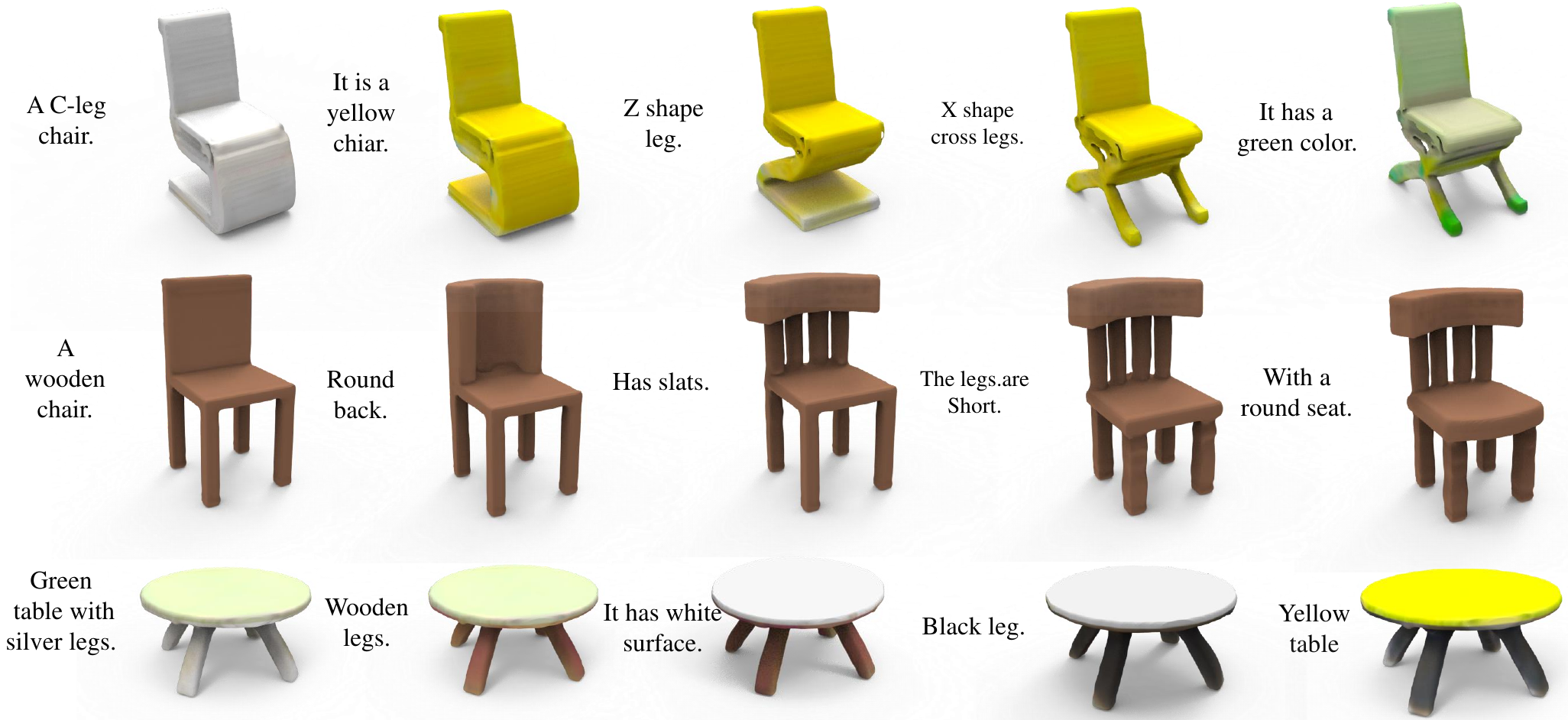}
\caption{Additional text-guided modification results. Our EXIM can effectively manipulate the interested region without affecting other regions. 
}
\label{fig:manipulation}
\vspace*{-2.75mm}
\end{figure*}

\begin{figure*}
\centering
\includegraphics[width=0.99\textwidth]{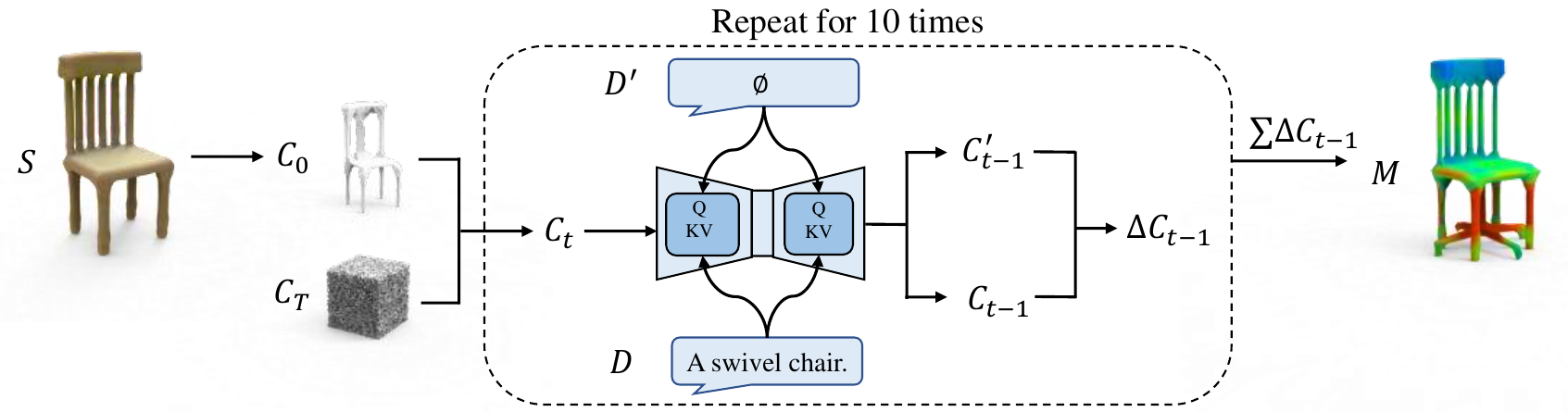}
\caption{The pipeline of deriving the mask using DiffEdit~\cite{couairon2022diffedit}.}
\label{fig:diffedit}
\vspace*{-2.75mm}
\end{figure*}

\section{Details on the Mask Prediction}


To perform local modification, we first generate a mask $M$ to identify the region of interest and guide subsequent modification operations. We adopt the recent DiffEdit approach~\cite{couairon2022diffedit}, which focuses on 2D image editing tasks, and adapt it to our 3D shape modification problem.

To identify the region of interest described by the text $D$, this approach exploits the differences in diffusion model predictions conditioned on two distinct texts, $D$ and $D^\prime$ (an empty text). The pipeline for deriving $M$ is depicted in Figure~\ref{fig:diffedit}. Given a shape $S$, we initially transform it into the wavelet domain, introduce noise to a predefined timestamp $t$ according to Equation (1) in the main paper, and obtain the noisy wavelet volume $C_t$. We then denoise $C_t$ to $C_{t-1}$ by conditioning the diffusion model on the input text $D$ and an empty text $D^\prime$, respectively. Given $C_t$, these two text prompts result in different noise estimations: $C_{t-1}$ and $C_{t-1}^\prime$. For instance, in the case illustrated in Figure~\ref{fig:diffedit}, $D$ aims to create a swivel chair, while the empty text prompt $D^\prime$ seeks to reconstruct the original shape with four legs $S$. In other regions, such as the back and seat, both $D$ and $D^\prime$ tend to predict similar noise to reconstruct the original shape $S$, as these areas are less relevant to the text prompt ``a swivel chair''.  Utilizing this observation, we can estimate the region of interest by calculating the difference between $C_{t-1}$ and $C_{t-1}^\prime$, denoted as $\Delta C_{t-1}$. We repeat this process ten times and normalize the summation of $\Delta C_{t-1}$ to values between 0 and 1. Subsequently, we can derive the binary mask $M$ using thresholding. As demonstrated in Figure~\ref{fig:diffedit}, we can effectively and automatically locate both the leg regions of the given shape $S$ and the swivel region for generating new structures, which are initially empty in $S$.

\section{Analysis on the Pseudo Annotations}

\subsection{Pseudo Annotation Creation}

In existing work, TAPS3D~\cite{wei2023taps3d} aims to generate pseudo captions for 3D shapes using CLIP~\cite{radford2021learning}. However, this approach requires the manual collection of ShapeNet-related nouns to create a vocabulary set, which significantly restricts the richness of the resulting descriptions. Furthermore, as explored in~\cite{liu2023iss}, there exists a noticeable gap between CLIP's text and image features, leading to potential inconsistencies between CLIP-based attribute retrieval and the actual 3D shape. As a result, the descriptions generated for ShapeNet chairs using this method contain an average of only $2.84$ words, which is insufficient for providing a detailed description of a shape.

\begin{figure}
\centering
\includegraphics[width=0.99\columnwidth]{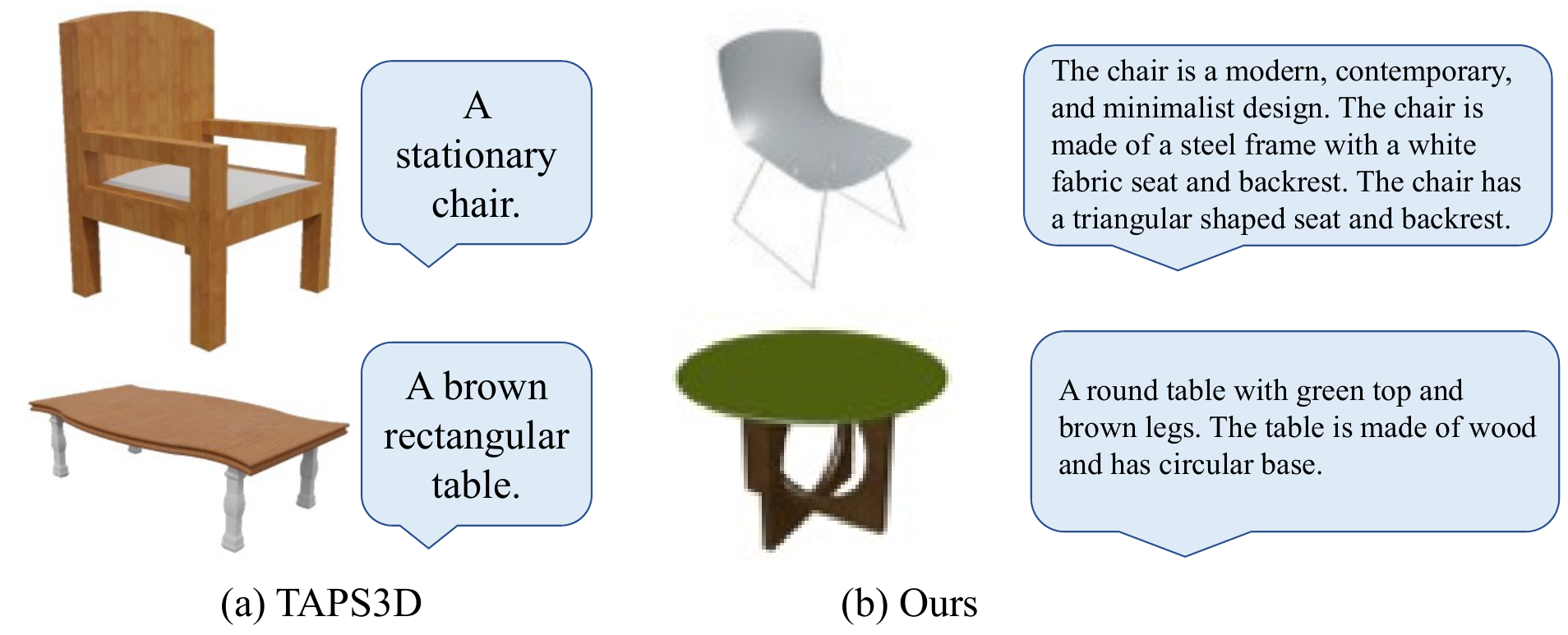}
\caption{(a) TAPS3D~\cite{wei2023taps3d} can only create simple captions while (b) our pseudo annotations contain rich details.}
\label{fig:blip}
\vspace*{-2.75mm}
\end{figure}

To address the aforementioned issues, we employ the large-scale vision-language model BLIP-v2~\cite{li2023blip} to generate descriptions for rendered images of 3D shapes using the prompt, "Question: please describe the [category] in as much detail as possible, including its color, shape, structure, parts, and attributes. Answer: ". As demonstrated in Figure~\ref{fig:blip}, our method is capable of producing detailed descriptions for the 3D shapes, allowing us to extend our approach to a wider range of categories beyond chairs and tables. 


\subsection{Performance of Pseudo Annotations}\label{sec:pseudo}

To evaluate the effectiveness of the pseudo-annotations and examine the potential of our approach for application to a broader range of categories, we train our method using texts generated by the image captioning model BLIP-v2~\cite{li2023blip}. We perform an analysis on the chair category from ShapeNet~\cite{shapenet2015} in order to compare our results with those obtained using human-annotated texts.

\begin{figure}
\centering
\includegraphics[width=0.99\columnwidth]{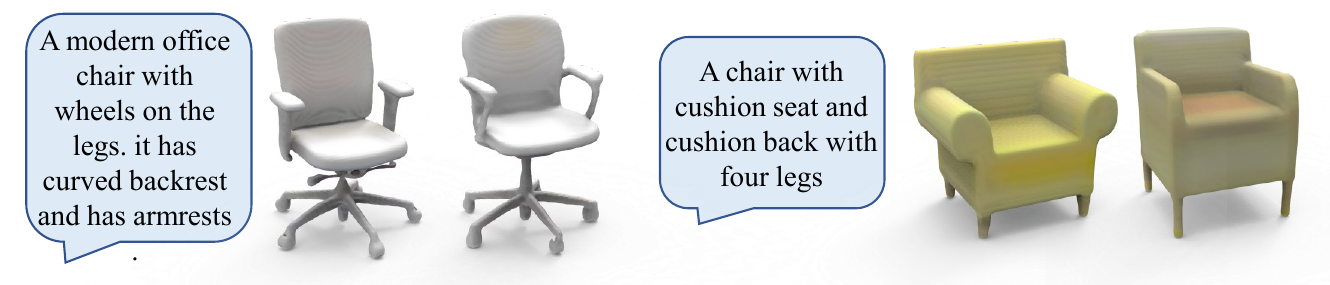}
\caption{Qualitative evaluation on the pseudo annotation quality. Left: trained with pseudo annotation. Right: trained with human-annotated data.  Our model can produce plausible shapes consistent with the human-annotated text even trained with pseudo annotations generated by the image captioning model BLIP-v2~\cite{li2023blip}. }
\label{fig:ablation_text}
\vspace*{-2.75mm}
\end{figure}

Figure~\ref{fig:ablation_text} and Table~\ref{tab:ablation_text} indicate that both of our models trained on pseudo-annotations achieve performance comparable to those trained on human-annotated texts. Specifically, the model trained using pseudo-annotations is slightly inferior to the one trained with human-annotated texts in terms of text-shape consistency and shape generation quality. However, our pseudo-annotations outperform the annotated ones regarding color generation. This can be partially attributed to the fact that nearly all pseudo-texts contain color descriptions, whereas annotated texts may not always include them.


\begin{table}[t]
	\centering
		\caption{Quantitative analysis on our pseudo annotations. }
	\resizebox{.9\columnwidth}{!}{
		\begin{tabular}{cccc}
			\toprule[1pt]
                        Method &   CLIP-S ($\uparrow$)   & FPD  ($\downarrow$) & FID ($\downarrow$)  \\ \hline
Pseudo annotation & 60.21  & 291.16 & \textbf{35.5} \\
Human annotation & \textbf{71.75}  & \textbf{192.17} &  36.8 \\
			\bottomrule[1pt]
	\end{tabular}}
    \vspace*{-1mm}
\label{tab:ablation_text}
\end{table}

\section{Comparison with Wavelet Representation}

To further demonstrate the generation capability of EXIM, we create a baseline following the state-of-the-art 3D shape generation approach~\cite{hui2022neural} which adopts wavelet representation. Specifically, given the coarse 3D shapes $S$ generated by our Explicit Diffusion Stage, we adopt an additional high-frequency wavelet domain component $H$ besides $C_0$ to predict details on $S$ following~\cite{hui2022neural} instead of our Implicit Refinement Stage. 

\begin{figure}
\centering
\includegraphics[width=0.99\columnwidth]{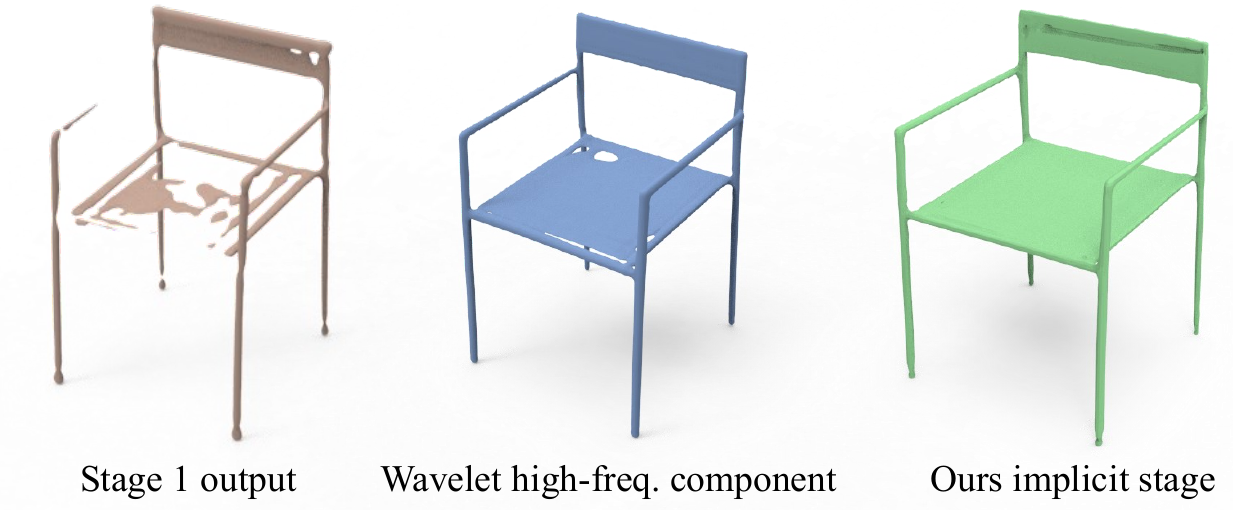}
\caption{Based on the output of our explicit stage (brown), our implicit stage (green) can better recover high-frequency details while keeping the original topology compared with the wavelet high-frequent components adopted in~\cite{hui2022neural} (blue).  }
\label{fig:wavelet}
\vspace*{-2.75mm}
\end{figure}

As shown in Figure~\ref{fig:wavelet}, our stage-2 refinement result (green) can better recover high-frequency details (see the holes) on the seat while maintaining the original structures compared with the wavelet high-frequent components adopted in~\cite{hui2022neural}. Further, 
the quantitative results shown in Table~\ref{tab:wavelet} demonstrate our approach outperforms our baseline consistently in all metrics. Beyond that, our implicit stage can produce plausible colors while the wavelet baseline cannot. 

\begin{table}[t]
	\centering
		\caption{Ablation study on hybrid representation. }
	\resizebox{0.99\linewidth}{!}{
		\begin{tabular}{c|cccc}
			\toprule[1pt]
                        \multirow{2}*{Method} & 
                          \multicolumn{2}{c}{COV ($\uparrow$)} & \multicolumn{2}{c}{MMD ($\downarrow$)} \\
                         & CD         & EMD        & CD         & EMD  \\ \hline
Stage 1
& 39.8          & 38.7 & 14.0          & 16.3    \\ \hline 
wavelet high freq.~\cite{hui2022neural}           
& 40.3    & 39.8 & 13.9 & 16.1 \\ \hline 
EXIM stage 2 & 
\textbf{41.7} & \textbf{42.4} & \textbf{13.8} & \textbf{15.9} \\ \hline 
			\bottomrule[1pt]
	\end{tabular}}
    \vspace*{-1mm}
\label{tab:wavelet}
\end{table}

\begin{figure}
\centering
\includegraphics[width=0.99\columnwidth]{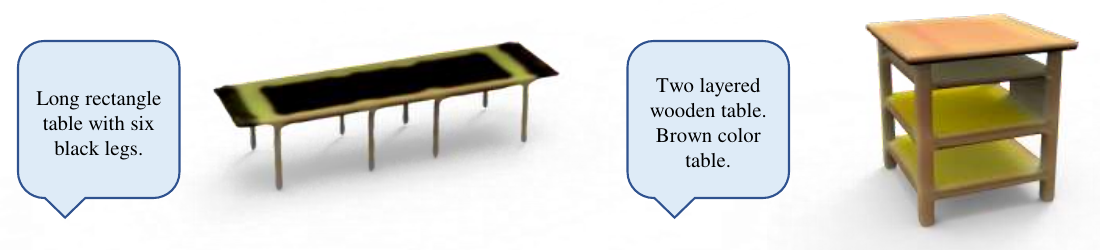}
\caption{Our generative results may not always be consistent with the text description, e.g., the numerical numbers. }
\label{fig:limitation}
\vspace*{-2.75mm}
\end{figure}

\section{Details on TSDF and 3D Wavelet Transform}

In this section, we provide additional information on TSDF and wavelet transform as suggested in \cite{hui2022neural}. 

The Truncated Signed Distance Function (TSDF) is a mathematical function used to quantify the distance from a query point $p$ to the surface of a 3D shape. This distance possesses a sign, denoting positivity for points exterior to the object and negativity for those within the object's boundaries. The TSDF undergoes truncation, implying that it's calculated within a limited distance threshold from the object's surface. In this work, we follow~\cite{hui2022neural} to normalize each shape to be in the [-0.45, +0.45] range and perform grid-based calculations for SDF values with a resolution of $256^3$~\cite{wang2022dual}. Then we truncate the distance values in the range of [-0.1, +0.1] to get the TSDF $S$.

To obtain the low-frequency wavelet component of the TSDF $S$ of the size $256^3$, we decompose its TSDF volume into a high-frequency wavelet volume $H^1$ and a low-frequency wavelet volume $C^1$ using the biorthogonal wavelets filter~\cite{cohen1992biorthogonal} and repeat the above process for $J$ times to obtain the low-frequency wavelet component $C^J$  for conducting the diffusion process.
Note that the above process is reversible: given a  low-frequency wavelet component $C^J$, we can reconstruct the shape $\hat{S}$ with details corresponding to  components of higher frequencies $\{H^1, C^1, C^2, ..., C^{J-1}\}$ discarded. Following~\cite{hui2022neural}, 
we choose $J$ to be $3$, resulting in a wavelet volume $C^3$ with size $46^3$ and perform diffusion process in this compact wavelet space. 

Readers may refer to \cite{hui2022neural} for more details.

\section{Limitations}

Despite the superior performance of our approach, there are still some limitations to this work. 
First, the performance of the categories trained on pseudo-annotations is largely constrained by the image captioning model, such as BLIP-v2~\cite{li2023blip}. Although our analysis in Section~\ref{sec:pseudo} of this supplementary file demonstrates the satisfactory quality of our pseudo-annotations, their quality still falls short compared to human-annotated ones. However, it is important to note that our approach is orthogonal to advancements in image captioning models, and we could adopt more recent models, such as miniGPT4~\cite{zhu2023minigpt}, to generate higher-quality pseudo-annotations. Additionally, utilizing large-scale image-language models for automatic annotation is a promising research direction.
Second, in text-guided 3D shape modification, automatically deriving an accurate mask $M$ from a text prompt remains an unsolved problem. Although we can employ DiffEdit~\cite{couairon2022diffedit} to locate the region of interest, there is no guarantee that the approach will always precisely segment the desired region. Therefore, we provide a secondary option for users to manually refine the mask or directly select the area of interest. While this manual strategy is a common choice in many existing works~\cite{nichol2021glide,lugmayr2022repaint}, automatically locating the region of interest would be a more desirable solution.
Third, the generated shapes may not always be consistent with the input text. For example, as shown in Figure~\ref{fig:limitation}, our model creates a table with eight legs instead of the mentioned six and produces a four-layered table when it should have been two layers.
Also, while image captioning enables training the generative model without paired text-shape data, our method still requires large-scale, category-specific 3D data for captioning. This poses a limitation compared to zero-shot methods like DreamFusion~\cite{poole2022dreamfusion}.
At last, although our approach can generate plausible colors on the 3D shapes in a feed-forward manner, it's still difficult for our current implementation to generate high-fidelity textures like existing methods that can generate rich color and textures often rely on a time-consuming per-scene optimization, leveraging 2D priors, e.g., DreamFusion~\cite{poole2022dreamfusion}.  

\bibliographystyle{ACM-Reference-Format}
\bibliography{bibliography}

\end{document}